\documentclass[10pt,journal,compsoc]{IEEEtran}
\usepackage{subcaption}
\usepackage{multirow}
\usepackage{graphicx}
\usepackage{amsmath}
\usepackage{amssymb}
\usepackage{algorithm}
\usepackage{algpseudocode}
\usepackage{xcolor}
\usepackage[normalem]{ulem}
\graphicspath{ {./images/} }
\ifCLASSOPTIONcompsoc
  \usepackage[nocompress]{cite}
\else
  \usepackage{cite}
\fi
\ifCLASSINFOpdf
\else
\fi
\begin{document}
\title{FLrce: Resource-Efficient Federated Learning with Early-Stopping Strategy}

\author{Ziru~Niu,
        Hai~Dong,~\IEEEmembership{Senior~Member,~IEEE},
        A. K. ~Qin,~\IEEEmembership{Senior~Member,~IEEE},
        and~Tao~Gu,~\IEEEmembership{Fellow,~IEEE}
\IEEEcompsocitemizethanks{\IEEEcompsocthanksitem Ziru Niu and Hai Dong are with the School of Computing Technologies, RMIT University, Melbourne, VIC 3000, Australia.\protect\\
E-mail: ziru.niu@student.rmit.edu.au, hai.dong@rmit.edu.au.
\IEEEcompsocthanksitem A. K. Qin is with the Department of Computing Technologies, Swinburne University of Technology, Hawthorn, VIC 3122, Australia.\protect\\ E-mail: kqin@swin.edu.au.
\IEEEcompsocthanksitem Tao Gu is with the Department of Computing, Macquarie University, Sydney, New South Wales, Australia, \protect\\ E-mail: tao.gu@mq.edu.au
}
\thanks{Manuscript accepted August 14, 2024. (Corresponding author: Hai Dong)\par DOI:10.1109/TMC.2024.3447000}}
\markboth{IEEE Transactions on Mobile Computing, 2024}%
{Shell \MakeLowercase{\textit{et al.}}: Bare Demo of IEEEtran.cls for Computer Society Journals}

\IEEEtitleabstractindextext{%
\begin{abstract}
Federated Learning (FL) achieves great popularity in the Internet of Things (IoT) as a powerful interface to offer intelligent services to customers while maintaining data privacy. Under the orchestration of a server, edge devices (also called clients in FL) collaboratively train a global deep-learning model without sharing any local data. Nevertheless, the unequal training contributions among clients have made FL vulnerable, as clients with heavily biased datasets can easily compromise FL by sending malicious or heavily biased parameter updates. Furthermore, the resource shortage issue of the network also becomes a bottleneck. Due to overwhelming computation overheads generated by training deep-learning models on edge devices, and significant communication overheads for transmitting deep-learning models across the network, enormous amounts of resources are consumed in the FL process.
This encompasses computation resources like energy and communication resources like bandwidth. To comprehensively address these challenges, in this paper, we present FLrce, an efficient FL framework with a \textbf{r}elationship-based \textbf{c}lient selection and \textbf{e}arly-stopping strategy. FLrce accelerates the FL process by selecting clients with more significant effects, enabling the global model to converge to a high accuracy in fewer rounds. FLrce also leverages an early stopping mechanism that terminates FL in advance to save communication and computation resources. Experiment results show that, compared with existing efficient FL frameworks, FLrce improves the computation and communication efficiency by at least 30\% and 43\% respectively.  
\end{abstract}

}

\maketitle

\IEEEdisplaynontitleabstractindextext

\IEEEpeerreviewmaketitle

\IEEEraisesectionheading{\section{Introduction}\label{sec:introduction}}

\IEEEPARstart{W}{ith} the rapid growth of the Internet of Things (IoT), edge devices have garnered extensive utilization in efficiently capturing and retaining vast volumes of data at the periphery of a network. To process these data and offer intelligent services to customers, a deep-learning model is usually required \cite{lecun2015deep}. However, traditional centralized machine learning approaches that require edge devices to upload data to a central server for further processing, are no longer applicable, as the process of data transmission significantly increases the risk of data leakage and violates user privacy. Owing to privacy concerns, it is desired that edge devices process these highly privacy-sensitive data locally rather than upload them to a remote server \cite{FLspm}. To this end, federated learning (FL) emerges. FL is a distributed machine learning paradigm that allows edge devices to collaboratively train a global model without sharing any private data \cite{fedavg}. In FL, devices (also named \textit{clients}) train a machine learning model locally from their personal datasets, and exchange the model parameters with the server. Compared with raw data, the risk of data leakage for transmitting the parameters becomes much smaller \cite{fedavg, fieldguide, fliotsurvey}. This capability enables FL to train a universal global model that can provide high-quality services to end users in a wide range of privacy-sensitive applications, such as voice assisting \cite{flvoice}, face recognition \cite{fedface}, human activity recognition \cite{fedhar} and healthcare \cite{health_survey}. \par
\begin{figure}
    \centering
    \includegraphics[scale=0.7]{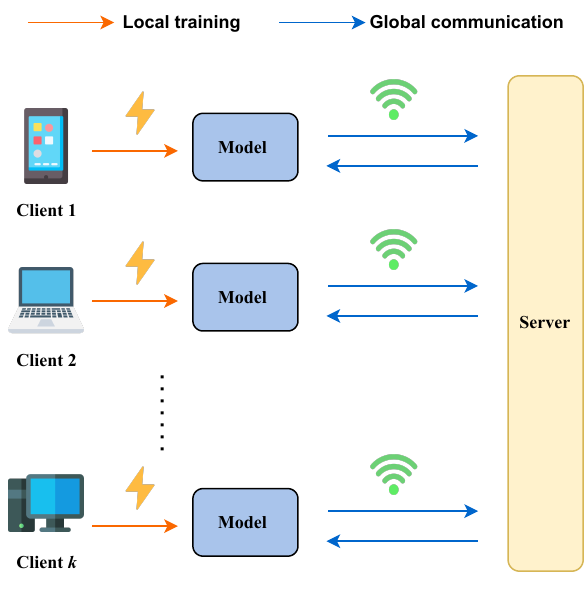}
    \caption{FL faces the dilemma of expensive computation and communication costs in IoT networks.}
    \label{shortage}
\end{figure}
Even though FL remarkably upholds data privacy, when it comes to application, the effectiveness of FL is usually compromised by the issues of resource shortage and client heterogeneity in real-world IoT networks. On one hand, \emph{due to the resource constraints of edge devices, the \textit{communication and computation resources} of an IoT network are usually finite, presenting a critical challenge to address the resource shortage issue of FL in IoT settings} \cite{flRCsurvey, mocha, FLspm}. As shown in Figure \ref{shortage}, first, an IoT network with limited bandwidth support can hardly withstand the exorbitant communication cost in FL when deep-learning models with massive amounts of parameters are communicated frequently between clients and the server \cite{pa, FLspm, flstc}.
Second, due to the significant energy consumption of training deep learning models on edge devices \cite{autofl}, FL usually generates high computation costs which can be unaffordable for IoT systems with deficient energy supply. To solve the resource shortage problem, an efficient FL framework that can reduce the computation and communication costs without degrading accuracy is expected. \par
\begin{figure}
    \centering
    \includegraphics[scale=0.7]{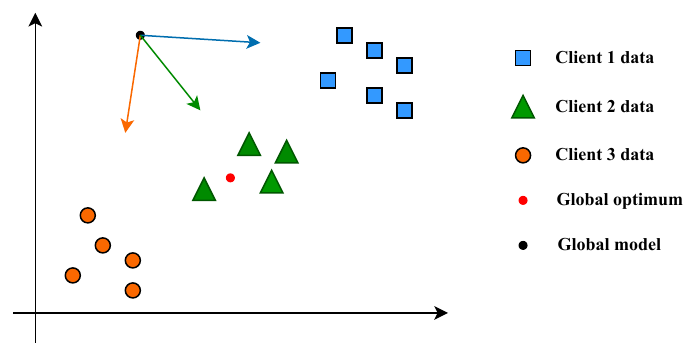}
    \caption{A simple 2-dimensional demonstration of unequal training contributions among FL clients.}
    \label{noniiddatademo}
\end{figure}
On the other hand, \emph{the geographically distributed edge devices naturally collect data that are non-identical independent (non-iid), leading to unbalanced datasets and heterogeneous training contributions among clients} \cite{fltrust, pa}, as shown by Figure \ref{noniiddatademo}. Specifically, some clients are more \emph{important} than others, as their local data distributions resemble the global data distribution (e.g. client 2 in Figure \ref{noniiddatademo}), while some clients are non-important or even harmful to FL because of their severely biased local data distributions (e.g. clients 1 and 3 in Figure \ref{noniiddatademo}). To overcome this challenge, a robust FL framework that can maintain accuracy with heterogeneous client contributions is desired. \par
Several works have been proposed to overcome the resource shortage issue of FL, including \emph{message compression}, \emph{accuracy relaxation} and \emph{dropout}. Message compression techniques \cite{flstc, fedcom, flexcom, terngrad, 8bit, qsgd} propose that clients locally compress the parameter updates before sending them to the server, so as to save communication resources. However, both client heterogeneity and the computation resource shortage issue are neglected in these works. Accuracy relaxation methods \cite{fedprox, fedparl, pyramid} allow clients to reduce the number of local training epochs, thereby saving computation resources with the expense of slight accuracy sacrifice. However, these works do not consider saving communication resources or dealing with client heterogeneity in FL. Dropout techniques \cite{feddrop, randdrop, fjord, hermes} let clients train a "sub-model", which is a subset of the global model parameters. Compared with the origin model, the computation and communication overheads for training and transmitting a sub-model become smaller. Therefore, both the computation and communication resources can be saved in these works. However, client heterogeneity is still not considered in dropout. \par
Studies have been conducted to tackle the client heterogeneity problem, which can generally be divided into two groups. The first group of works \cite{fldetecrtor, fedcbs, shieldfl} propose to identify the non-important or malicious clients based on the parameter updates they provide, non-important or malicious clients are less likely to be selected by the server to participate in FL. The second group of works \cite{fltrust, adbfl, robustfed} propose to alleviate the negative impacts of non-important or malicious clients by assigning smaller weights to the parameter updates of these clients in the aggregation phase. However, the bottleneck of resource shortage in FL is neglected in these works. \par
Motivated by the limitation of the state-of-the-art, in this paper, we present \textbf{FLrce}, i.e. \textbf{F}ederated \textbf{L}earning with \textbf{r}elationship-based \textbf{c}lient selection and \textbf{e}arly stopping strategy, to simultaneously address the client heterogeneity and resource shortage problems of FL in IoT settings. \par
Firstly, to overcome the challenge of client heterogeneity, FLrce leverages an innovative client selection strategy that tends to select the clients that are more "important" to improve the robustness and the learning efficiency of FL. The importance of a client is the sum of its relationship degree with other clients, where the relationship degree between two clients is computed by the cosine similarity or orthogonal distance between their parameter updates. This client selection strategy enables the global model to reach a high accuracy in fewer rounds, which greatly improves the performance of FL with heterogeneous client contributions. \par
Secondly, to overcome the challenge of resource shortage, FLrce utilizes an early stopping mechanism that terminates FL in advance to reduce the total rounds of training. We notice that when the global model "oscillates" between clients’ local optima, it becomes very hard for the model to reach the global optimal solution, making FL extremely inefficient. Accordingly, FLrce detects the oscillation between the clients’ parameter updates using cosine similarity, when a large degree of oscillation happens, i.e. the number of conflicting updates exceeds a pre-defined threshold, the early stopping mechanism will be triggered. Subsequently, the overall resource consumption can be significantly reduced, including computation resources (e.g. energy) and communication resources (e.g. bandwidth). The main contributions of our work are summarized as follows:
\begin{itemize}
    \item We design a relationship-based client selection strategy that significantly increases the robustness of FL and helps FL maintain accuracy with unbalanced contributions from heterogeneous clients.
    \item We implement an early-stopping strategy to prevent redundant consumption of the computation and communication resources with very little (sometimes even zero) accuracy sacrifice. To the best of our knowledge, our work is the first one to apply the early stopping technique \cite{esbutwhen} to reduce resource consumption in FL.
    \item We implement FLrce on a physical platform and evaluate FLrce on real-world datasets. The experiment results show that, compared with existing efficient FL frameworks, FLrce at least increases the efficiency of the computation/communication resource utilization by 30\% and 43\% respectively on average. 
\end{itemize}
The structure of this paper is organized as follows. Section \ref{BK} introduces the background and motivation. Section \ref{DOF} presents the implementation of the proposed method FLrce. Section \ref{EXP} exhibits the experimental results and provides critical analysis. Section \ref{RW} briefly introduces the related work. Section \ref{conclusion} summarizes this article and lists future directions. \par

\section{Background and Motivation} \label{BK}
\subsection{Federated Learning}
Within an IoT network, there exist a central server \(S\) and \(M\) edge devices. A device \(k\), \(k \in \{1,2,...,M\}\) is referred as a \emph{client} in FL, named as \(C_{k}\) . The universal set of all clients \(\mathbb{C}\) is defined as \(\mathbb{C}=\{C_{k}: 1 \leq k \leq M \}\). Each \(C_{k}\) has a local dataset \(D_{k}\) containing \(n_{k}\) samples: \(D_{k} = \{(\boldsymbol{x}_{i},y_{i}), 1\leq i \leq n_{k}, i \in \mathbb{Z}\}\), with \(\boldsymbol{x}_{i}\) and \(y_{i}\) being the features and the label of the \(i\)-th sample in \(D_{k}\). Totally, there are \(n = \sum_{k=1}^{M}n_{k} \) samples across all local datasets. Both the server and the clients store a machine learning model for any possible classification or regression tasks. The server stores a global model \(w\), and each \(C_{k}\) stores a local model \(w_{k}\). Besides, we use \(w^{*}_{k}\) to denote the optimal model for client \(C_{k}\), which is defined as the model with the best performance on \(C_{k}\)'s local dataset. That is:
\begin{equation}\label{eq:loptmodel}
w^{*}_{k} = \mathop{\arg\min}_{w}F_{k}(w) \triangleq \mathop{\arg\min}_{w} \frac{1}{n_{k}} \sum_{i=1}^{n_{k}} l((\boldsymbol{x}_{i}, y_{i}), w)
\end{equation}
\(F_{k}\) here is the local objective function, which is equivalent to the empirical risk over dataset \(D_{k}\), with \(l\) being the classification/regression loss on a single sample. \par

We use \(w^{*}\) to denote the optimal global model, which is defined as the model with the best average performance across all local datasets, that is:
\begin{equation}\label{eq:goptmodel}
w^{*} = \mathop{\arg\min}_{w}F(w) \triangleq \frac{1}{M}\sum_{k=1}^{M}F_{k}(w)
\end{equation}
\(F\) here is the global objective function by minimizing which we attempt to find \(w^{*}\), it is defined as the mean of all clients' local objective functions. \par

Let \(T\) be the maximum number of global iterations in FL, and \(P\) be the number of active clients that participate in FL at each iteration. At round \(t \in \{1,2,...,T\}\), the server selects a set of active clients \(\mathbb{C}_{t}\) (\(\mathbb{C}_{t} \subset \mathbb{C}, |{\mathbb{C}_{t}}|=P\)). The server then sends the current global model \(w^{t}\) to all clients in \(\mathbb{C}_{t}\). In parallel, each \(C_{k}\) (\(C_{k} \in \mathbb{C}_{t}\)) loads the received parameter into its local model and computes the parameter update \(u^{t}_{k}\) through local optimization:
\begin{equation}\label{eq:lopt}
    \begin{split}
    & w_{k} \gets w^{t}\\
    & u^{t}_{k} = - \eta\nabla F_{k}(w_{k}) \\
    \end{split}
\end{equation}
where \(\eta\) is the learning rate and \(\nabla F_{k}(w_{k})\) is the gradient of local objective function \(F_{k}\) with respect to $w_{k}$. Then \(C_{k}\) uploads the parameter update \(u_{k}^{t} \) to the server. The server aggregates the received parameter updates using weighted average and updates the global model as Equation (\ref{wa}) shows:
\begin{equation}\label{wa}
\begin{split}
    & w^{t+1} = w^{t} + \sum_{k: \ C_{k} \in \mathbb{C}_{t}}p_{k}u_{k}^{t} \\
    & p_{k} = \frac{n_{k}}{\sum_{k': \ C_{k'} \in \mathbb{C}_{t}}n_{k'}} \\
\end{split}
\end{equation}

\begin{figure}
    \centering
    \includegraphics[scale=0.7]{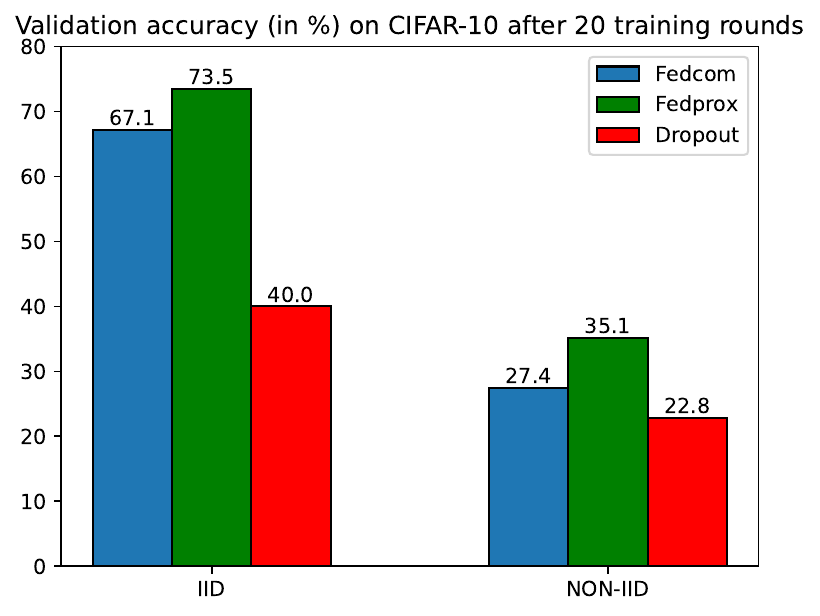}
    \caption{Existing efficient FL methods easily become vulnerable to non-iid client data.
    }
    \label{moti1}
\end{figure}
\begin{figure}[ht]
    \centering
    \begin{subfigure}[b]{0.4\textwidth}
        \centering
        \includegraphics[width=\textwidth]{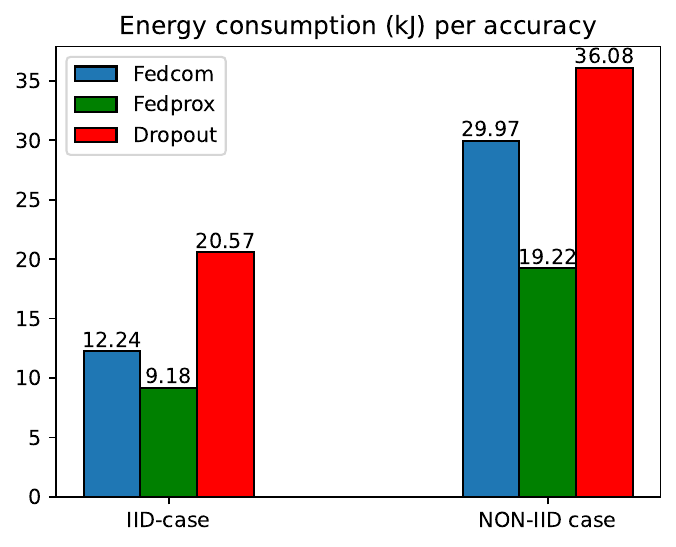}
        \caption{Required energy consumption (kJ) per accuracy.}
        \label{img:moti2_energy}
    \end{subfigure}
     \begin{subfigure}[b]{0.4\textwidth}
        \centering
        \includegraphics[width=\textwidth]{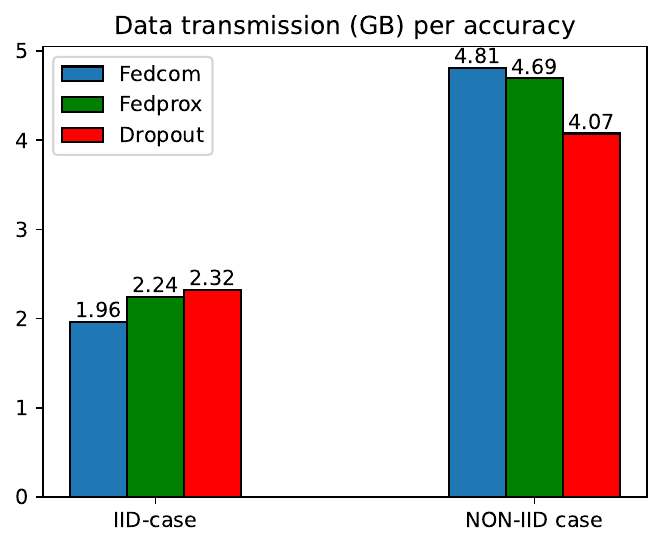}
        \caption{Required data transmission (GB) per accuracy.}
        \label{img:moti2_bd}
    \end{subfigure}
    \caption{Due to the vulnerability to heterogeneous client contributions, existing efficient FL methods consume more communication and computation resources to reach the same target accuracy in the non-iid case.
    }
    \label{moti2}
\end{figure}
\begin{figure*}
    \centering
    \includegraphics[scale=0.6]{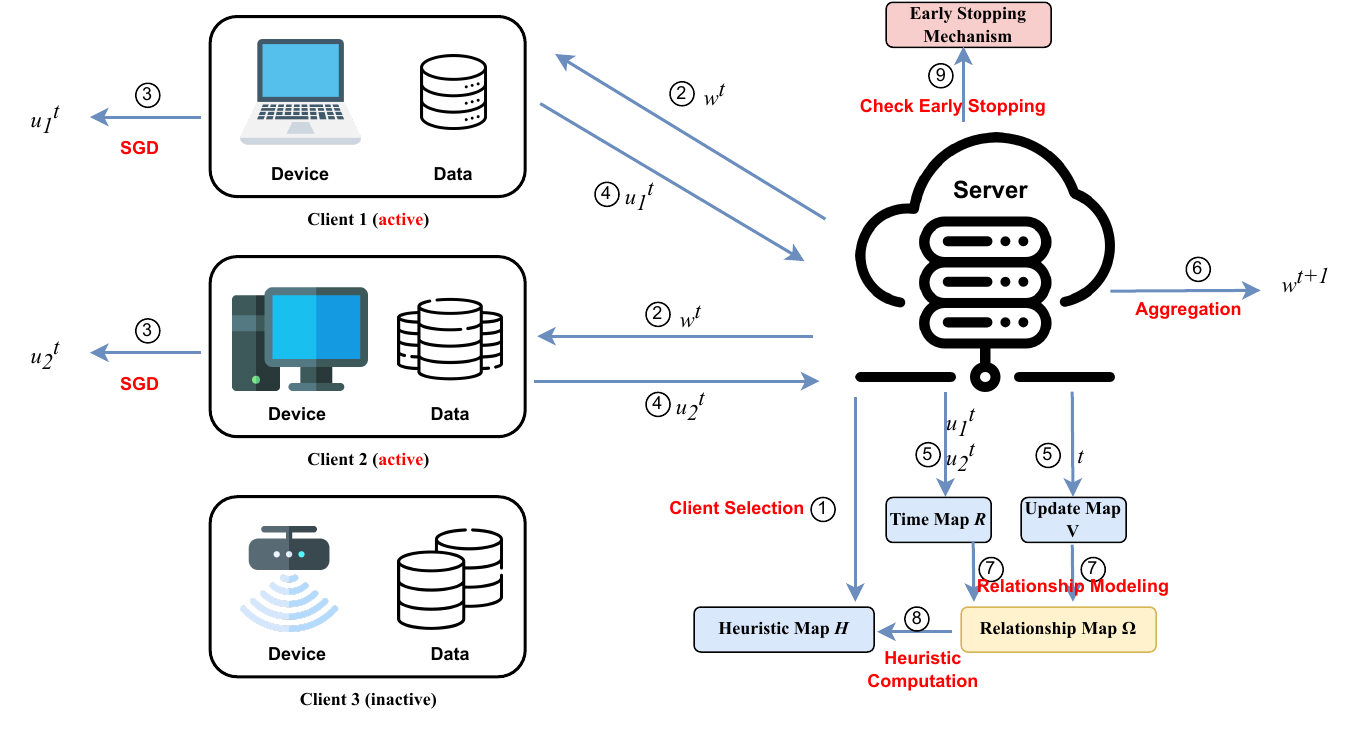}
    \caption{Overview of the FLrce framework.}
    \label{flrceflow}
\end{figure*}
\subsection{Degraded Performance Caused by Client Heterogeneity}
Due to the non-iid local data distribution of FL clients, the performance of existing efficient FL frameworks can be easily compromised by unbalanced training contributions from heterogeneous clients. For illustration, we implement Fedcom \cite{fedcom}, Fedprox \cite{fedprox}, and Dropout \cite{randdrop} on the CIFAR10 dataset \cite{cifar}, and compare their performances between the cases where the local data distributions are iid and non-iid. In the iid case, all clients share a centralized global dataset. In the non-iid case, data are unevenly distributed to clients following the experimental setting of \cite{nofear}. The simulation runs on 100 clients with 20 iterations, and other parameter settings can be found in the experiment section. As shown in Figure 3, existing efficient FL methods are significantly impeded by client heterogeneity in the non-iid case. The reason is that, these methods usually employ a trade-off strategy over local clients to reduce the computation or communication cost, such as model pruning \cite{randdrop}, training epoch reduction \cite{fedprox} and message compression \cite{fedcom}. While for important clients, applying these strategies might prohibit FL from learning sufficient knowledge from the important local data, and consequently degrades the global model accuracy.
\subsection{Additional Resource Consumption Resulting from Low Learning Efficiency}
Owing to the degraded accuracy caused by client heterogeneity, the expected computation cost (energy) and communication cost (data transmission) per accuracy of existing efficient FL methods inevitably escalate as Figure \ref{moti2} shows. Consequently, given any target accuracy, these methods are expected to consume more computation and communication resources to reach that accuracy. Specifically, Fedcom, Fedprox and Dropout respectively consume 2.45×, 2.09×, 1.75× computation and communication resources to achieve the same target accuracy compared with the iid-case. 
\section{Design of FLrce} \label{DOF}
In this section, we introduce the proposed FLrce framework. Specifically, subsection \ref{FLrce_overview} lays out a high-level demonstration of how FLrce works. Subsections \ref{FLrce_cs} and \ref{FLrce_es} respectively explain the client selection strategy and the early stopping mechanism of FLrce. Subsection \ref{FLrce_all} summarizes all the details and presents a comprehensive overview of FLrce. 
\subsection{Solution Overview} \label{FLrce_overview}
A high-level workflow of FLrce is presented in Figure \ref{flrceflow}. \textcircled{1}: At the beginning of round \(t\), the server decides the set of active clients based on a client selection strategy \(h\). \(h\) takes \(H\) as input and returns the server a set of clients \(\mathbb{C}_{t}\), where \(H\) is the heuristic map that stores the heuristic values of clients (see Algorithm \ref{functionh}). 
\textcircled{2}: The server broadcasts the current global model \(w^{t}\) to all selected clients. 
\textcircled{3}: In parallel, each participating client \(C_{k}\) (\(C_{k} \in \mathbb{C}_{t}\)) updates the received parameter by applying stochastic gradient descent (SGD) over its personal dataset (see Equation (\ref{eq:lopt})).
\textcircled{4}: Each \(C_{k}\) uploads the parameter updates \(u^{t}_{k}\) to the server. 
\textcircled{5}: For each \(C_{k} \in \mathbb{C}_{t}\), the server writes the corresponding \(u^{t}_{k}\) to \(V_{k}\), and writes \(t\) to \(R_{k}\). \(V\) is an update map that stores each client's latest update, \(R\) is a time map that stores each client's last active round. 
\textcircled{6}: The server aggregates the received updates and updates the global model parameter (see Equation (\ref{wa})). 
\textcircled{7}: The server evaluates the relationship between each selected \(C_{k}\) and other clients, and writes the relationships into a relationship map \(\Omega \in \mathbb{R}^{M\times M}\), using a relationship modeling function \(g\) (see Algorithm \ref{functiong}). 
\textcircled{8}: The server computes each \(C_{k}\)'s latest heuristic value based on the newest \(\Omega\) and updates the heuristic map \(H\) (see Equation (\ref{heuristic})). 
\textcircled{9}: The server examines whether the early-stopping mechanism has been triggered. If so, the server terminates FL prematurely (see subsection \ref{FLrce_es}). 
If not, the server moves ahead to round \(t+1\) and restarts from step \textcircled{1}. The summary of all mathematical notations is listed in Table \ref{tab:notations}. \par
\begin{table}[ht]
    \centering
    \begin{tabular}{c|c}
         \hline
         \textbf{Notations} & \textbf{Semantics} \\
         \hline
         \(S\) & Central server \\
         \(C_{k}\) & The \(k\)-th client\\
         \(M\) & Total number of clients \\
         \(T\) & Maximum number of global iterations \\
         \(t\) & The \(t\)-th round of global training \\
         \(P\) & Number of active clients per round \\
         \(\mathbb{C}\) & Universal set of clients \\
         \(\mathbb{C}_{t}\) & Set of active clients at round \(t\) \\
         \(D_{k}\) & The local dataset of \(C_{k}\) \\
         \(n_{k}\) & Number of samples in \(D_{k}\) \\
         \(n\) & Total number of samples in all local datasets \\
         \(w^{t}\) & Global model at round \(t\) \\
         \(w_{k}^{t}\) & \(C_{k}\)'s local model at round \(t\) \\
         \(w^{*}\) & Optimal global model \\
         \(w^{*}_{k}\) & Optimal model for \(C_{k}\) \\
         \(F\) & Global objective function \\
         \(F_{k}\) & Local objective function for \(C_{k}\) \\
         \(\nabla F_{k}(w_{k})\) & Gradient of \(F_{k}\) with respect of \(w_{k}\) \\
         \(\eta\) & Learning rate \\
         \(u_{k}^{t}\) & Parameter update of \(C_{k}\) at round \(t\) \\
         \(H\) & Map of clients' heuristic values \\
         \(R\) & Map of clients' latest active rounds \\
         \(V\) & Map of clients' latest updates \\
         \(\Omega\) & Map of clients' pairwise relationships \\
         \(g\) & Relationship modeling function \\
         \(h\) & Client selection strategy \\
         \(RM\) & Relationship Modeling \\
         \(ES\) & Early stopping mechanism \\
         \(\phi\) & Explore-exploit factor of client selection \\
         \(\psi\) & Early stopping threshold \\
         \hline
    \end{tabular}
    \caption{Summary of Notations}
    \label{tab:notations}
\end{table}
\subsection{Relationship-based Client Selection}\label{FLrce_cs}
Motivated by existing works that address client heterogeneity by avoiding selecting non-important clients \cite{fedcbs,fldetecrtor}, FLrce presents a novel client selection strategy that tends to select clients who are more important to FL to address client heterogeneity and promote accuracy. This design enables the global model to reach a high accuracy with fewer training rounds and less resource consumption. An important client helps improve the global model's mean accuracy and will maintain a positive relationship with most of the others. Therefore, FLrce utilizes a relationship modeling (RM) algorithm that estimates clients' importance to FL by observing their pairwise relationships. The RM algorithm consists of two components: \textit{synchronous RM} and \textit{asynchronous RM}. \par
\begin{figure}
    \centering
    \includegraphics[scale=0.7]{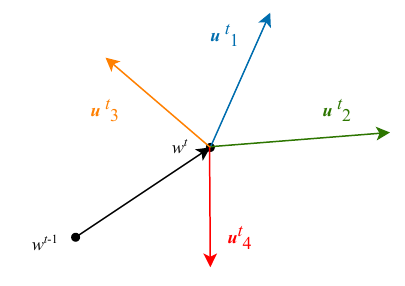}
    \caption{Server receives and compares four updates at round \emph{t}.}
    \label{sim_relation_fig}
\end{figure}

\textbf{Synchronous Relationship Modeling with Cosine Similarity}. Motivated by \cite{fltrust, clusterfl}, we model the relationship between clients who are simultaneously selected by computing the \emph{cosine similarity }(\emph{cossim}) between their parameter updates. For example, suppose at round \(t\), the server receives the updates from clients 1\(\sim\)4 as shown by Figure \ref{sim_relation_fig}. In this case, the server computes the relationship degree between every client pair by Equation (\ref{sim_relation_eq}):
\begin{equation}\label{sim_relation_eq}
\forall p,q \in \{1,2,3,4\}, p \neq q: \Omega_{p,q} \gets cossim(u^{t}_{p}, u^{t}_{q})
\end{equation}
From Figure \ref{sim_relation_fig} and Equation (\ref{sim_relation_eq}), we can see that client 1 has positive relationships with clients 2 and 3, but clients 2 and 3 conflict with each other, and client 4 is negatively correlated with all others. \par
\begin{figure}
    \centering
    \includegraphics[scale=0.7]{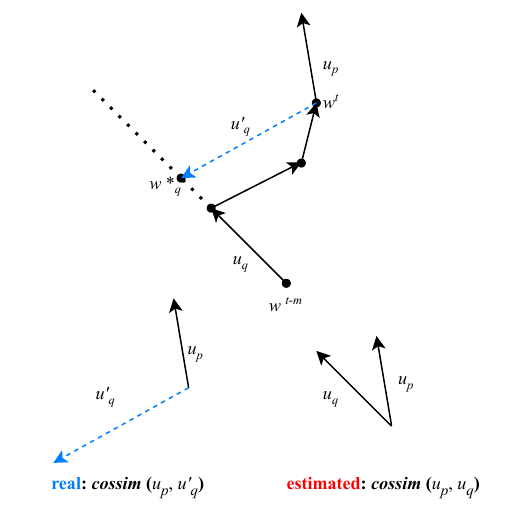}
    \caption{Asynchronous relationship modeling with cosine similarity can diverge drastically from the real value.}
    \label{coserror}
\end{figure}

Even though cosine similarity has been proven to work well in estimating synchronous clients' relationships \cite{fltrust}, if it is used between asynchronous clients, the result might be inaccurate. For example, in Figure \ref{coserror}, the server selects \(C_{q}\) at round \(t-m \ (m > 1)\) and selects \(C_{p}\) at round \(t\), and the corresponding updates are \(u_{q}\) and \(u_{p}\) respectively. Using cosine similarity, the server would get an inaccurate estimation \(cossim(u_{p}, u_{q})\) that diverges drastically from the real value \(cossim(u_{p}, u'_{q})\), where \(u'_{q}\) is the vector pointing from \(w^{t}\) towards the underlying local optimum \(w_{q}^{*}\), as Figure \ref{coserror} shows. \par
When the participation rate of clients is high, the server receives a large amount of updates each round, synchronous RM will be sufficient. For example, given 100 FL clients with a participation rate of 50\%, the expected rounds that the server takes to learn a comprehensive relationship map \(\Omega\) is 4 (\(100^{2}/50^{2}\)). However, within a real-world network, the number of participating clients in each communication round is often limited \cite{FLspm, flscale}, making synchronous RM inefficient. For example, for 100 FL clients with a participation rate of 10\%, the server would require 100 (\(100^{2}/10^{2}\)) rounds by the expectation to learn \(\Omega\) completely. Therefore, it is preferable to design an asynchronous RM strategy that empowers the server to retrieve the relationships faster. \par 
\begin{figure}
    \centering
    \begin{subfigure}[b]{0.20\textwidth}
        \centering
        \includegraphics[width=\textwidth, height=0.16\textheight]{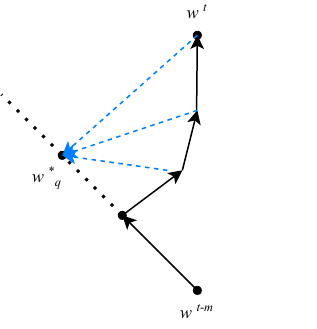}
        \caption{}
        \label{om1point}
    \end{subfigure}
    \hfill
    \begin{subfigure}[b]{0.20\textwidth}
        \centering
        \includegraphics[width=\textwidth,
        height=0.16\textheight]{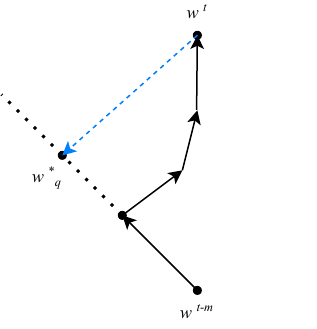}
        \caption{}
        \label{om4points}
    \end{subfigure}
    \hfill
    \begin{subfigure}[b]{0.20\textwidth}
        \centering
        \includegraphics[width=\textwidth,
        height=0.16\textheight]{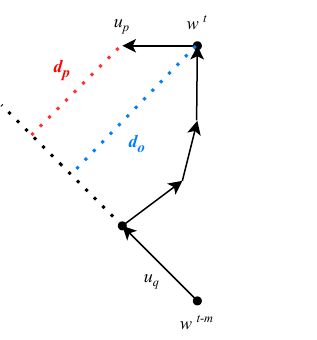}
        \caption{}
        \label{omtriangle}
    \end{subfigure}
    \caption{A demo. why FLrce uses orthogonal distance for asynchronous RM: as the orthogonal distance between \(w^{t}\) and \(u_{q}\) increases, the global model is expected to diverge further away from \(C_{q}\)'s local optimal solution.}
    \label{rmom}
\end{figure}

\textbf{Asynchronous Relationship Modeling with Orthogonal Distance}. FLrce conducts asynchronous RM using \emph{orthogonal distance} (\emph{orthdist}), with Figure \ref{rmom} demonstrating how it works. First, suppose the server receives an update \(u_{q}\) from \(C_{q}\) at round \(t-m\). Then \(C_{q}\)'s local optimal solution \(w_{q}^{*}\) \textit{approximately} lies on the direction that \(u_{q}\) points to, as Figure \ref{om1point} shows. As a matter of fact, due to the randomness of the local optimization procedure (like SGD), \(u_{q}\) does not necessarily point to \(w_{q}^{*}\), while we assume it does for simplicity. Second, given \(w_{q}^{*}\) and the global model parameter \(w^{t}\), the Euclidean distance \(\parallel w^{t}-w_{q}^{*}\parallel_{2}\) can be used to represent how far the current global model is towards \(C_{q}\)'s local optimal solution. For example, in Figure \ref{om1point}, \(\parallel w^{t}-w_{q}^{*}\parallel_{2}\) is the length of the blue dashed arrow. However, due to data privacy in FL, the coordinates of \(w_{q}^{*}\) are unknown and computing \(\parallel w^{t}-w_{q}^{*}\parallel_{2}\) is prohibitive. \par

\begin{algorithm}
\caption{Relationship modeling function \(g\)}
\label{functiong}
\begin{algorithmic}[1]
\Require update \(u^{t}_{k}\), old map \(\Omega_{k}\), update map \(V\), time map \(R\), round \(t\)
\For {\(j \in \{1,...,M\}, j \neq k\)}
\State compare \(u_{k}^{t}\) with other clients' updates:
\State \(v \gets V_{j}, t' \gets R_{j}\) 
\If{$t' \geq t-1 $}
\State synchronous relationship modeling:
\State \hspace{5mm} \(\Omega_{k,j} \gets cossim(v, u^{t}_{k}) \) 
\Else
\State asynchronous relationship modeling:
\State \hspace{5mm} \(\Omega_{k,j} \gets \mathop{\max}(1 - \frac{orthdist(w^{t} + u^{t}_{k}, v)}{orthdist(w^{t}, v)}, -1)\) 
\EndIf
\EndFor
\State \Return new map \(\Omega_{k}\)
\end{algorithmic}
\end{algorithm}
\begin{algorithm}
\caption{Client selection strategy \(h\)}
\label{functionh}
\begin{algorithmic}[1]
\Require heuristic map \(H\), \emph{explore-exploit} factor \(\phi\), number of participants \(P\), available clients \(\mathbb{C}\)
\State determine if \emph{explore} or \emph{exploit} based on \(\phi\)
\If{\emph{exploit}}
\State select clients with the largest heuristic values:
\State \hspace{5mm}\(\mathbb{C}' \gets\) sorted(\(\mathbb{C}\), key=\(H\), reverse=\textbf{True})
\State \hspace{5mm}\(\mathbb{C}_{t} \gets\) The first \(P\) elements in \(\mathbb{C}'\)
\Else
\State randomly explore clients:
\State \hspace{5mm}\(\mathbb{C}_{t} \gets\) random \(P\) samples from \(\mathbb{C}\)
\EndIf
\State \Return \(\mathbb{C}_{t}\)
\end{algorithmic}
\end{algorithm}
Even though we cannot directly know the distance between \(w^{t}\) and \(w_{q}^{*}\), we can estimate the change of the distance through the orthogonal distance of \(w^{t}\) towards \(u_{q}\). Concretely, when the orthogonal distance between \(w^{t}\) and \(u_{q}\) increases, the global model is more likely to leave \(w_{q}^{*}\). Similarly, when the orthogonal distance decreases, the global model is more likely to approach \(w_{q}^{*}\). For example, in Figure \ref{om4points}, when the orthogonal distance between \(w^{t}\) and \(u_{q}\) increases, \(w^{t}\) is expected to diverge further away from \(w_{q}^{*}\). \par
Based on this point, suppose the server receives an update \(u_{p}\) from \(C_{p}\) at round \(t+1\), to evaluate the relationship degree between \(C_{p}\) and \(C_{q}\),  we mainly observe the change of the orthogonal distance between \(u_{p}\) and \(w^{t}\) after incorporating \(C_{p}\)'s an update, as Figure \ref{omtriangle} shows. Therefore, the relationship degree between \(C_{p}\) and \(C_{q}\) can be calculated by Equation (\ref{od_relation_eq}): \par
\begin{equation}\label{od_relation_eq}
\begin{split}
    & \Omega_{p,q} \gets \mathop{\max}(1-d_{p}/d_{o}, -1) \\
    & = \mathop{\max}(1 - \frac{orthdist(w^{t} + u^{t}_{p}, u^{t-m}_{q})}{orthdist(w^{t}, u^{t-m}_{q})}, -1) \\
\end{split}
\end{equation}
\(d_{p} > d_{o}\) indicates that the global model is possibly moving away from \(C_{q}\)'s local optimum after incorporating the update from \(C_{p}\), then a negative relationship degree will be assigned between \(C_{p}\) and \(C_{q}\). If \(d_{p} < d_{o}\), \(C_{p}\) and \(C_{q}\) will have a positive relationship degree for a similar reason. To control the relationship degree on the same scale as cosine similarity, we make the result in Equation (\ref{od_relation_eq}) be no less than -1. The computation of Equation (\ref{od_relation_eq}) will not infringe on data privacy, as it only requires clients' parameter updates and the global model parameter. 
With the synchronous and asynchronous RM combined, the RM function \(g\) is implemented in Algorithm \ref{functiong}. \par
\textbf{Heuristic Values Computation.} The heuristic value of a client \(C_{k}\) reflects the importance of \(C_{k}\) to the collaborative learning process. Motivated by \cite{fltrust}, FLrce evaluates the importance of a client as the combination of its relationships with other peers. Therefore, we define the heuristic of \(C_{k}\) as a summation of its relationship degrees with other clients:
\begin{equation}\label{heuristic}
    H_{k} = \sum_{j=1, j\neq k} ^ {M} \Omega_{k,j}
\end{equation}
Furthermore, the client selection strategy \(h\) is implemented in Algorithm \ref{functionh}. \(h\) is defined as an \emph{explore-exploit} algorithm, for exploration, it randomly selects \(P\) clients. For exploitation, it selects clients with the top-\(P\) heuristic values. The \emph{explore-exploit} factor \(\phi\) depends on \(t\). In earlier rounds, the server knows little about clients' relationships and is more likely to explore. In later rounds, with sufficient knowledge about the relationships, the server has a higher chance to exploit. 
\begin{figure}
    \centering
    \includegraphics[scale=0.8]{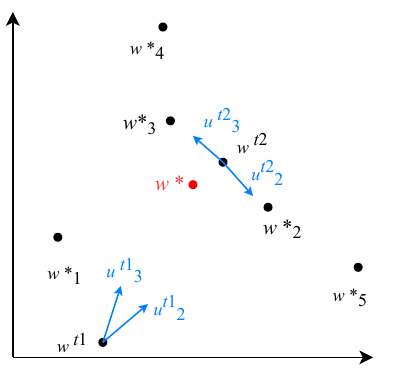}
    \caption{When a large degree of conflict occurs among the most important clients, the ES mechanism will be triggered.}
    \label{esdemo}
\end{figure}
\begin{algorithm}
\caption{Early stopping criterion \(ES\)}
\label{esc}
\begin{algorithmic}[1]
\Require round \(t\), participating clients \(\mathbb{C}_{t}\), number of participants \(P\), threshold \(\psi\)
\If{\(t\) is an \emph{exploit} round}
\State \(conflicts \gets 0\)
\State calculate total conflicting pairs:
\For{\(k \in \{1,...,M\}, C_{k} \in \mathbb{C}_{t}\)}
  \For{\(C_{j} \in \mathbb{C}_{t}, j \neq k \)}
      \If{\(cossim(u_{k}^{t}, u_{j}^{t} ) < 0\)}
      \State \(conflicts \gets conflicts + 1\)
      \EndIf
  \EndFor
\EndFor
\State calculate average conflicting pairs:
\State \hspace{5mm} \(conflicts \gets \frac{1}{P}\times conflicts\)
\If{\(conflicts \geq \psi\)} 
\State \Return \textbf{True}
\EndIf
\EndIf
\State \Return \textbf{False}
\end{algorithmic}
\end{algorithm}
\subsection{Early Stopping Mechanism} \label{FLrce_es}
FLrce presents a novel early stopping mechanism settings that terminates FL ahead of the maximum training round. Unlike \cite{esbutwhen} which uses early stopping to prevent overfitting in centralized machine learning, the early stopping mechanism in FLrce aims to reduce the overall communication and computation resource consumption in decentralized FL settings. The point is that, as FL goes on, the global model tends to "oscillate" between the important clients' local optima. In this case, it becomes much more difficult for the global model to reach the global optimum, which triggers the ES mechanism to avoid further consumption of communication and computation resources. \par
For illustration, Figure \ref{esdemo} shows a simple two-dimensional example. Suppose a network with five clients \(C_{1} \sim C_{5}\). \(C_{2}\) and \(C_{3}\) are the two most important clients as their local optima are closer to the global optimum than others. Given \(P=2\), the server tends to to select \(C_{2}\) and \(C_{3}\) in an exploit round. Suppose two exploit rounds \(t_{1}\) and \(t_{2}\). At \(t_{1}\), the global model \(w^{t_{1}}\) is far away from the optimal model \(w^{*}\), and the server finds \(C_{2}\) and \(C_{3}\) not conflicting as \(cossim(u_{2}^{t_{1}}, u_{3}^{t_{1}}) > 0\). On the contrary, \(w^{t_{2}}\) is closer to \(w^{*}\) and the server finds \(C_{2}\) and \(C_{3}\) conflicting as \(cossim(u_{2}^{t_{2}}, u_{3}^{t_{2}}) < 0\). In summary, selecting the important clients can accelerate training in earlier rounds. However, as FL goes on, important clients are likely to conflict with each other and provide inconsistent parameter updates. In this case, selecting the important clients is no longer effective in improving the accuracy of the global model. \par
\begin{algorithm}
\caption{FLrce}
\label{flrce}
\begin{algorithmic}[1]
\Require max iterations \(T\), number of participants \(P\), clients \(\mathbb{C}\), \textit{explore-exploit} factor \(\phi\), ES threshold \(\psi\)
    \State \textbf{Globally, server \(S\) executes}:
\State initialize \(w^{0}\), \(H\), \(\Omega\), \(R\), \(V\)
\For{\(t = 1,2,...,T\)}
\State client selection:
\State \hspace{5mm}  \(\mathbb{C}_{t} \gets h(H,T,P, \mathbb{C}, \phi)\) \Comment{Alg. \ref{functionh}}
\For{\(k \in \{1,...,M\}, C_{k} \in \mathbb{C}_{t}\)}
\State send \(w^{t}\) to \(C_{k}\)
\State receive \(u^{t}_{k}\) from \(C_{k}\)
\State write update/round to the update/time map:
\State \hspace{5mm} \(V_{k} \gets u^{t}_{k}, R_{k} \gets t\)
\EndFor
\State aggregation: 
\State \hspace{5mm} \(w^{t+1} \gets w^{t} + \sum_{k:C_{k} \in \mathbb{C}_{t}}p_{k}u_{k}^{t}\) \Comment{Eq. (\ref{wa})}
\For{\(k \in \{1,...,M\}, C_{k} \in \mathbb{C}_{t}\)}
\State update relationship:
\State \hspace{5mm} \(\Omega_{k} \gets g(u^{t}_{k}, V, R, t, \Omega_{k})\) \Comment{Alg. \ref{functiong}}
\State update heuristic:
\State \hspace{5mm} \(H_{k} = \sum_{j=1, j\neq k} ^ {M} \Omega_{k,j}\) \Comment{Eq. (\ref{heuristic})}
\EndFor
\State check early stopping:
\If{\(ES(t, \mathbb{C}_{t}, P, \psi) == \) \textbf{True}} \Comment{Alg. \ref{esc}} 
\State \textbf{break}
\EndIf
\EndFor
\State \Return \(w^{t+1}\)
\end{algorithmic}
\begin{algorithmic}
\State \textbf{Locally, each \(C_{k} \in \mathbb{C}_{t}\) in parallel executes}: 
\end{algorithmic}
\begin{algorithmic}[1]
    \State receive \(w^{t}\) from \(S\)
    \State update local model: 
    \State \hspace{5mm} \(w_{k} \gets w^{t}\)
    \State local training with SGD: 
    \State \hspace{5mm} \(u^{t}_{k} = - \eta\nabla F_{k}(w_{k})\) \Comment{Eq. (\ref{eq:lopt})}
    \State \textbf{return} \(u^{t}_{k}\) to \(S\)
\end{algorithmic}
\end{algorithm}
Furthermore, in the case of Figure \ref{esdemo}, after \(t_{2}\), instead of reaching \(w^{*}\), the global model will oscillate between \(w^{*}_{2}\) and \(w^{*}_{3}\), making it hard to further improve the mean accuracy. Therefore, we implement an ES mechanism to terminate FL in advance to avoid the unnecessary waste of communication and computation resources, as shown in Algorithm \ref{esc}. The variable \emph{conflicts} in Algorithm \ref{esc} is the degree of conflicts, representing the average number of conflicting peers that each selected client has. FLrce assumes that when \(conflicts\) exceeds a pre-defined threshold \(\psi\), the global model is probably oscillating between several local optima instead of moving towards the global optimum, in this case, the ES mechanism will be activated to avoid further resource consumption. For example, in Figure \ref{esdemo}, FLrce stops at \(t_{2}\) with \(conflicts=1\) and \(\psi=1\). The threshold \(\psi\) indicates the maximum degree that FLrce can tolerate the conflicts between the selected clients. The smaller \(\psi\) is, the easier the ES mechanism will be triggered. \par

\subsection{Put All the Pieces Together} \label{FLrce_all}
Combining all the aforementioned together, we present FLrce. The main components of FLrce are: \textbf{1.} The server evaluates clients' importance to FL after receiving their parameter updates. \textbf{2.} The server tends to select the most important clients to make the model quickly converge to a high accuracy. \textbf{3.} Training is stopped before the maximum iteration out of resource concerns. A comprehensive framework of FLrce is presented in Algorithm \ref{flrce}. The key factor that distinguishes FLrce from existing efficient FL methods is that existing works reduce the resource consumption by reducing the computation/communication cost per round, while FLrce reduce the overall resource consumption by reducing the total rounds of training. \par

\begin{table*}
    \centering
    \begin{tabular}{c|c|c|c|c|c|c|c|c}
        \hline
        \multirow{2}{4em}{\textbf{Datasets}} & \multirow{2}{4em}{\textbf{Classes}} & \multirow{2}{4em}{\textbf{Total samples}} & \textbf{Global iteration} & \textbf{Total clients} & \textbf{Active clients} & \multirow{2}{4em}{\textbf{Batch size}} & \multirow{2}{4em}{\textbf{Local epoch}} & \textbf{Learning rate} \\
        & & & \(T\) & \(M\) & \(P\) & & & \(\eta\)\\
        \hline
        EMNIST & 62 & 814,255 & \multirow{4}{4mm}{100} & \multirow{4}{4mm}{100} & \multirow{4}{4mm}{10} & 16 & \multirow{4}{4mm}{5} & 2e-4\\
        Google Speech & 35 & 101,012 & & & &  16 &  & 5e-4 \\
        CIFAR10 & 10 & 50,000 & & & & 128 &  & 0.1 \\
        CIFAR100 & 100 & 50,000 & & & & 128 &  & 0.5 \\
        \hline
    \end{tabular}
    \caption{Summary of the experiment settings}
    \label{tab:setting summary}
\end{table*}

\section{Experiment} \label{EXP}
In this section, we evaluate the performance of FLrce on several datasets to demonstrate its strength in saving computation and communication resources and maintaining accuracy with unbalanced training contributions from heterogeneous clients. The experiment code is available at: https://github.com/ZiruNiu0/FLrce. 
\begin{figure*}
    \centering
    \begin{subfigure}[b]{\textwidth}
        \centering
        \includegraphics[width=\textwidth]{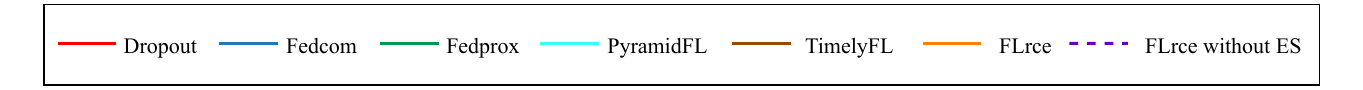}
    \end{subfigure}
    \begin{subfigure}[b]{0.24\textwidth}
        \centering
        \includegraphics[width=\textwidth, height=0.16\textheight]{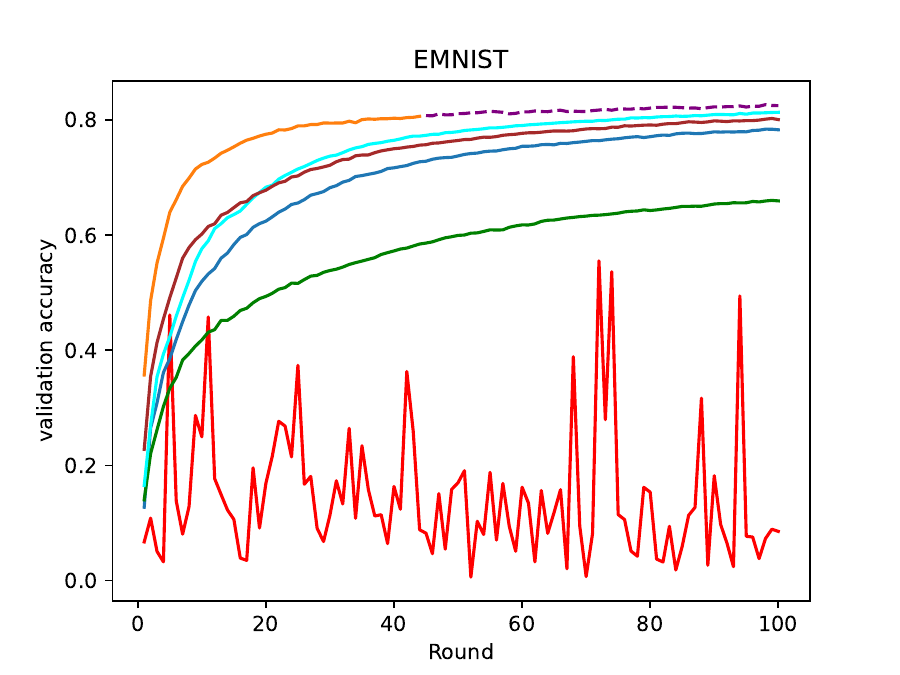}
        \caption{EMNIST}
        \label{emnistacc}
    \end{subfigure}
    \hfill
    \begin{subfigure}[b]{0.24\textwidth}
        \centering
        \includegraphics[width=\textwidth,
        height=0.16\textheight]{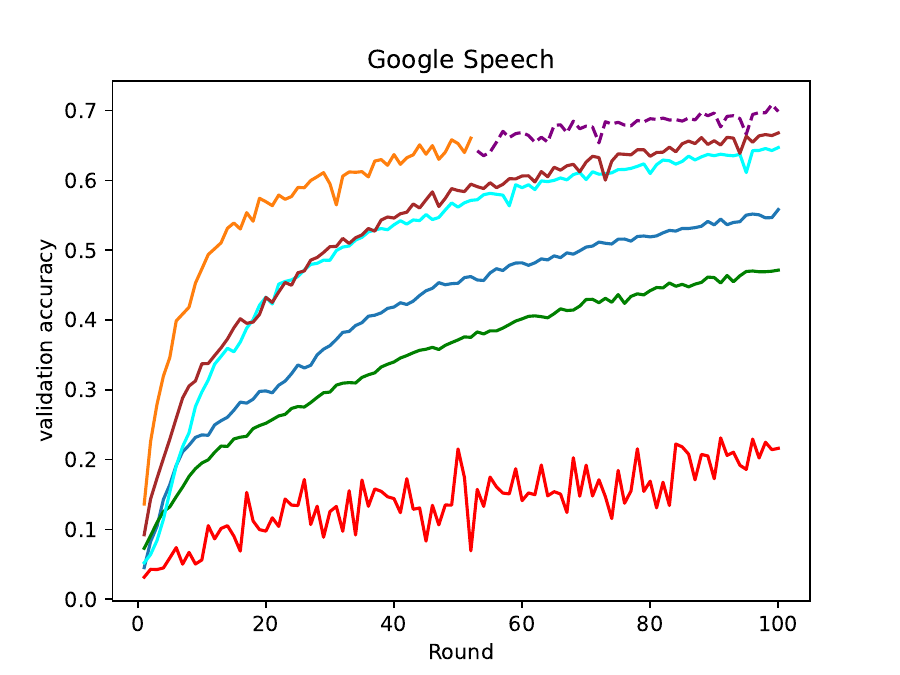}
        \caption{Google Speech}
        \label{voiceacc}
    \end{subfigure}
    \hfill
    \begin{subfigure}[b]{0.24\textwidth}
        \centering
        \includegraphics[width=\textwidth,
        height=0.16\textheight]{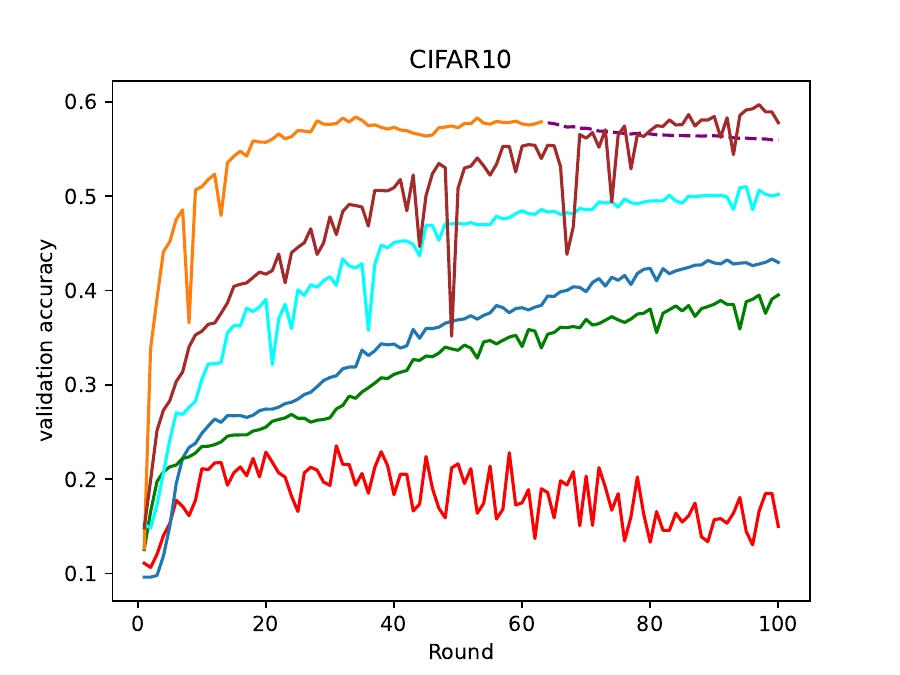}
        \caption{CIFAR10}
        \label{cifar10acc}
    \end{subfigure}
    \begin{subfigure}[b]{0.24\textwidth}
        \centering
        \includegraphics[width=\textwidth,
        height=0.16\textheight]{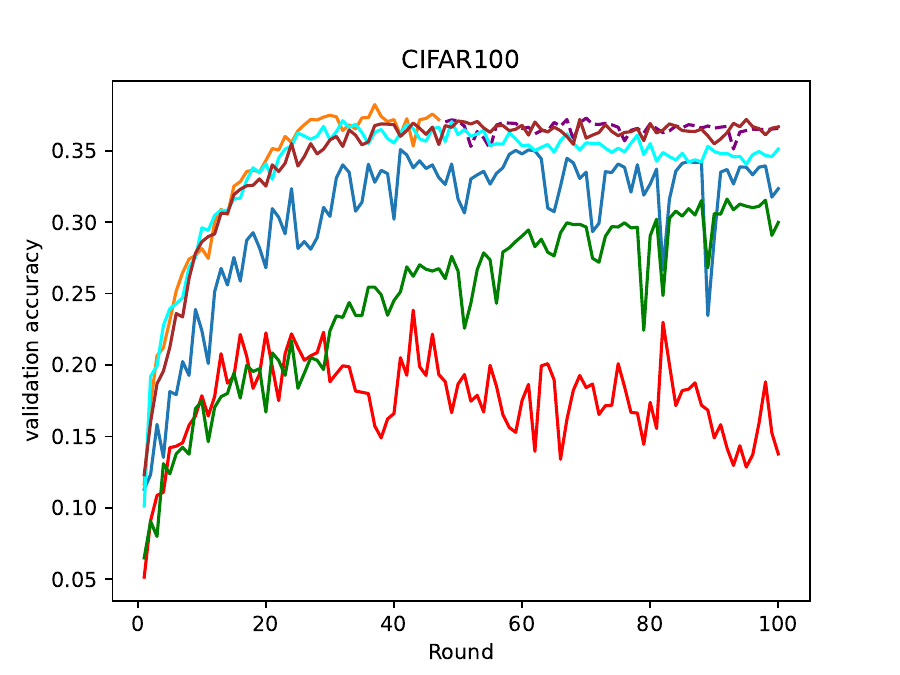}
        \caption{CIFAR100}
        \label{cifar100acc}
    \end{subfigure}
    \caption{Mean validation accuracy per round.}
    \label{img:acc}
\end{figure*}
\begin{table*}
    \centering
    \begin{tabular}{c|cc|cc|cc|cc|cc|cc}
    \hline
         \multirow{2}{10mm}{\textbf{Dataset}} & \multicolumn{2}{|c|}{\textbf{Fedcom}} & \multicolumn{2}{|c|}{\textbf{Fedprox}} & \multicolumn{2}{|c|}{\textbf{Dropout}} & \multicolumn{2}{|c|}{\textbf{PyramidFL}} & \multicolumn{2}{|c|}{\textbf{TimelyFL}} & \multicolumn{2}{c}{\textbf{FLrce}} \\
         &{\scriptsize Accuracy}&{\scriptsize Round}&{\scriptsize Accuracy}&{\scriptsize Round}&{\scriptsize Accuracy}&{\scriptsize Round}&{\scriptsize Accuracy}&{\scriptsize Round} &{\scriptsize Accuracy}&{\scriptsize Round} &{\scriptsize Accuracy}&{\scriptsize Round}\\
         \hline
         EMNIST & 78.2\% & 100 & 65.9\% & 100 & 8.5\% & 100 & 81.3\% & 100 & 80.06\% & 100 & 80.63\% & \textbf{44} \\
         Google Speech & 55.8\% & 100 & 47.1\% & 100 & 21.6\% & 100 & 64.7\% & 100 & 66.8\% & 100 & 66.05\% & \textbf{52} \\
         CIFAR10 & 42.9\% & 100 & 39.5\% & 100 & 14.9\% & 100 & 50.19\% & 100 & 57.8\% & 100 & 57.82\% & \textbf{63} \\
         CIFAR100 & 32.3\% & 100 & 29.9\% & 100 & 13.7\% & 100 & 35.11\% & 100 & 36.67\% & 100 & 37.57\% & \textbf{47} \\
         \hline
         Average & 52.3\% & 100 & 45.6\% & 100 & 14.6\% & 100 &  57.82\% & 100 & 60.33\% & 100 & 60.51\% & \textbf{52} \\ 
    \hline
    \end{tabular}
    \caption{Total rounds of training and final validation accuracy.}
    \label{tab:acc and round}
\end{table*}
\subsection{Experiment Setup}
\textbf{Datasets.} The experiment is conducted over several real-world datasets including: 
\begin{itemize}
    \item Extended MNIST (EMNIST) \cite{emnist} is an image dataset consisting of 814,255 handwritten characters from 3550 writers. Each sample is a black-and-white-based image with \(28 \times 28 \) pixels, belonging to one of the total 62 categories (10 digits and 52 upper/lower-case letters).  
    \item CIFAR10 \cite{cifar} is an image dataset consisting of 50,000 samples covering 10 classes of real-world objects. Each sample is a colorful image with \(32\times32\) pixels. 
     \item CIFAR100 \cite{cifar} is an image dataset containing 50,000 samples covering 100 classes of real-world objects. Each sample is a colorful image with \(32\times32\) pixels.
    \item Google Speech \cite{googlespeech} is a voice dataset consisting of 101,012 samples covering 35 human spoken words. Each sample is a short audio of a word pronounced by one of the total 2618 speakers. 
\end{itemize}
\textbf{Parameter settings.} For EMNIST and Google Speech, we use a convolutional neural network (CNN) with two convolutional layers and one fully-connected layer following \cite{randdrop}. For CIFAR10 and CIFAR100, we use a CNN with two convolutional layers and three fully-connected layers following \cite{hermes}. The number of global iterations is \(T=100\) and the number of clients' local training epochs is 5 universally following the settings in \cite{pyramid, fedcom}. The batch size is set to 128 for CIFAR10 and CIFAR100, and 16 for EMNIST and Google Speech by following \cite{fjord, pyramid}. We determine the learning rate \(\eta\) through hyperparameter tuning. If \(\eta\) is too large, all methods reach a high accuracy in the very first few rounds, making it extremely hard to distinguish their differences. If \(\eta\) is too small, the training goes slowly and requires much more iterations and time to produce a proper outcome. After hyperparameter tuning, the learning rate \(\eta\) is set to 2e-4 for EMNIST, 5e-4 for Google Speech, 0.1 for CIFAR10 and 0.5 for CIFAR100. Lastly, the explore-exploit factor \(\phi\) depends on \(t\), that is, the explore possibility of FLrce is 1.0 initially and decays by a coefficient of 0.98 every round following \cite{pyramid}. \par
\textbf{System implementation.} The experiment platform is built upon the Flower framework (1.4.0) \cite{flwr} and Pytorch 2.0.0 \cite{pytorch}. We build a virtual network consisting of one server and \(M=100\) clients, with \(P=10\)  clients participating in FL per round by following the experimental setting in \cite{randdrop}. The server runs on a remote cloud server and selects clients through the interface of Flower. Clients run the local training program on NVIDIA Jetson Nano Developer Kits \cite{nano}. We allow the edge devices to switch identities between several clients to conduct large-scale experiments with a limited number of physical devices \cite{fjord}.
\par

\textbf{Non-iid data distributions.} To create a training environment with heterogeneous client contributions, data are unevenly separated into clients following a Dirichlet distribution. Specifically, for EMNIST and Google Speech, data are allocated to clients based on the number of writers/speakers, with each client \(k\) accounting for \(p_{k}\) of the total 3550/2618 writers or speakers. The distributions of each \(p_{k}\) follow a Dirichlet Distribution \(Dir(\alpha)\) with a concentration parameter \(\alpha\). For CIFAR10 and CIFAR100, the proportions of samples from each class \(y\) vary among clients following a Dirichlet Distribution \(Dir_{y}(\alpha)\). The parameter \(\alpha\) of the Dirichlet Distribution is set to 0.1 following \cite{nofear}. \par

\textbf{Baselines.} As mentioned before, existing efficient FL methods can be divided into three categories: message compression, accuracy relaxation and dropout. Therefore, we compare FLrce against the typical works in these three categories. Out of resource concerns, we do not include any works that simply address client heterogeneity without taking efficiency into account. The baselines include:
\begin{itemize}
    \item \textit{Fedcom} \cite{fedcom} let clients compress the parameter updates before uploading the updates to the server so as to reduce communication costs.
    \item \textit{Fedprox} \cite{fedprox} allows clients to reduce the number of local training epochs to reduce the computation overheads and save computation resources.
    \item \textit{Dropout} \cite{randdrop} allows clients to train a "sub-model", which is a subset of the global model. Compared with the original model, the resource consumption for training and transmitting the sub-model becomes smaller.
    \item \textit{PyramidFL} \cite{pyramid} lets clients dynamically reduce local training epochs similar to Fedprox. Additionally, it applies a client selection strategy aiming to accelerate training by selecting clients with larger training losses and less training time.  
    \item \textit{TimelyFL} \cite{timelyfl} combines local epoch reduction and layer freezing altogether to reduce the computation cost of clients' local training tasks. 
\end{itemize}
\textbf{Metrics.} We comprehensively compare FLrce with the baselines in terms of accuracy, computation efficiency and communication efficiency. For accuracy, by the end of every global iteration, we test the global model on every local dataset and record the mean validation accuracy. For computation efficiency, we measure the overall energy consumption of the Jetson Nano Develop Kits during the local training process, and we combine the energy consumption with the final accuracy to derive the computation efficiency. For communication efficiency, we measure the bandwidth usage for transmitting the model between clients and the server, and we combine the bandwidth usage with the final accuracy to derive the communication efficiency. \par
\subsection{Experiment Results}
\subsubsection{Accuracy and Learning efficiency}
The experiment outcome shows that FLrce outperforms the baselines in terms of learning efficiency and final accuracy. From Figure \ref{img:acc}, we can see that the learning efficiency of FLrce is usually higher than the baselines, representing that FLrce can achieve higher accuracy given the same training rounds. Furthermore, with the help of the ES mechanism, FLrce is able to complete the global training task using only \(40\sim60\)\% of the maximum global iteration and still achieve equivalent or higher final accuracy. Specifically, as shown by Table \ref{tab:acc and round}, it takes only 44, 52, 63 and 47 rounds for FLrce to reach an accuracy of 80.06\%, 66.8\%, 57.8\% and 36.67\% on EMNIST, Google Speech, CIFAR10 and CIFAR 100 respectively. Comparatively, Fedcom uses 100 rounds to reach an accuracy of 78.2\%, 55.8\%, 42.9\% and 32.3\%. Fedprox uses 100 rounds to reach an accuracy of 65.9\%, 47.1\%, 39.5\% and 29.9\%. Dropout uses 100 rounds to reach an accuracy of 8.5\%, 21.6\%, 14.9\% and 13.7\%, which are completely exceeded by FLrce. Moreover, even though PyramidFL and TimelyFL achieve higher final accuracy than FLrce occasionally, they require much more training rounds and therefore more resources to reach that accuracy. Overall, even with fewer training rounds, FLrce still achieves the highest accuracy CIFAR10 and CIFAR100 and the second highest accuracy on EMNIST and Google Speech. On average, FLrce achieves the highest final accuracy with at least 0.18\% improvement compared with the baselines. \par
\begin{figure*}
    \begin{subfigure}[b]{0.24\textwidth}
        \centering
        \includegraphics[width=\textwidth,
        height=0.13\textheight]{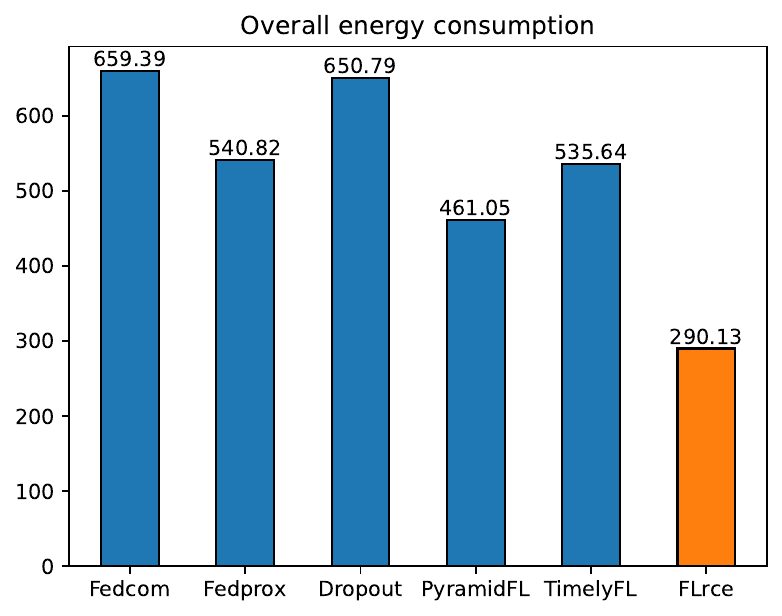}
        \caption{EMNIST}
        \label{emnistenergy}
    \end{subfigure}
    \begin{subfigure}[b]{0.24\textwidth}
        \centering
        \includegraphics[width=\textwidth,
        height=0.13\textheight]{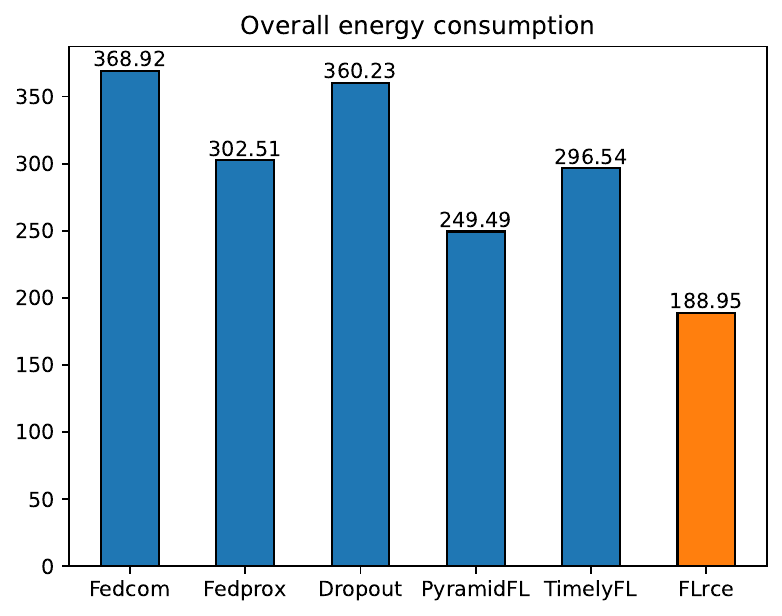}
        \caption{Google Speech}
        \label{voiceenergy}
    \end{subfigure}
    \begin{subfigure}[b]{0.24\textwidth}
        \centering
        \includegraphics[width=\textwidth,
        height=0.13\textheight]{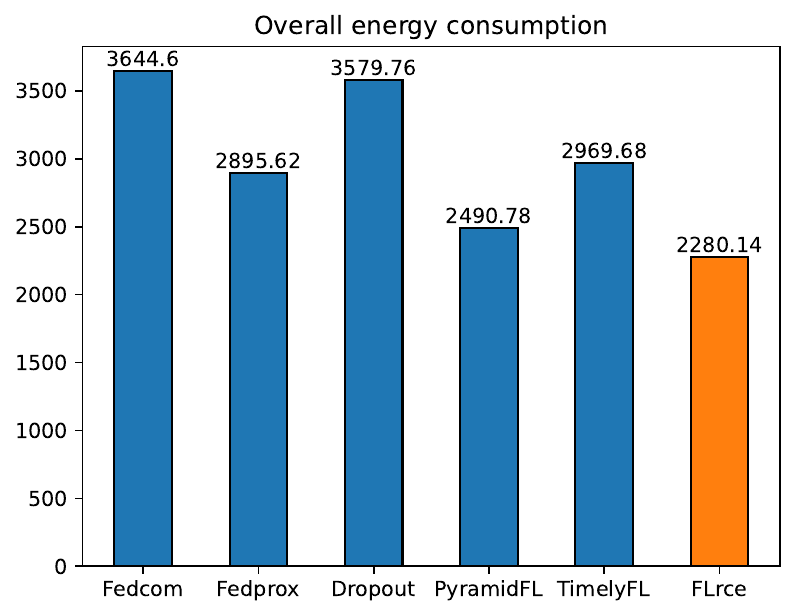}
        \caption{CIFAR10}
        \label{cifar10energy}
    \end{subfigure}
    \begin{subfigure}[b]{0.24\textwidth}
        \centering
        \includegraphics[width=\textwidth, height=0.13\textheight]{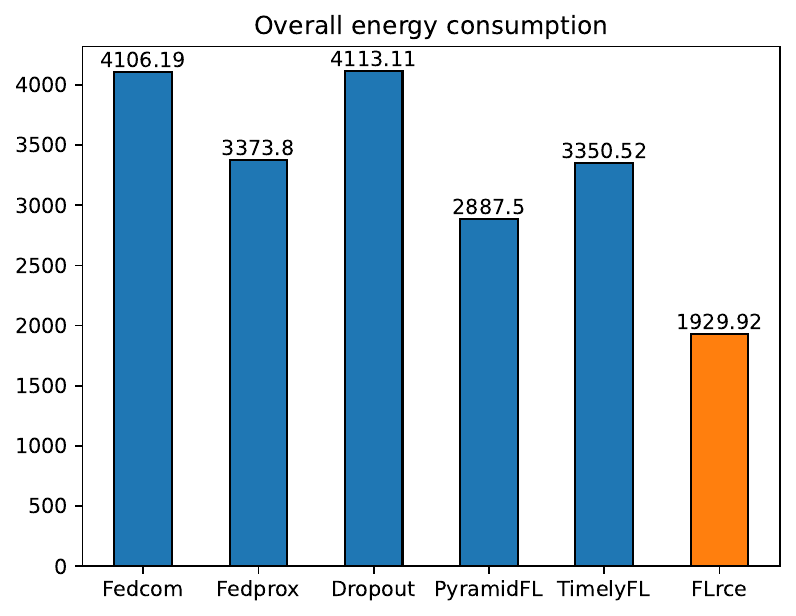}
        \caption{CIFAR100}
        \label{cifar100energy}
    \end{subfigure}
    \caption{Overall energy consumption (kJ).}
    \label{img:energy}
\end{figure*}
\begin{figure*}
    \centering
    \includegraphics[width=\textwidth]{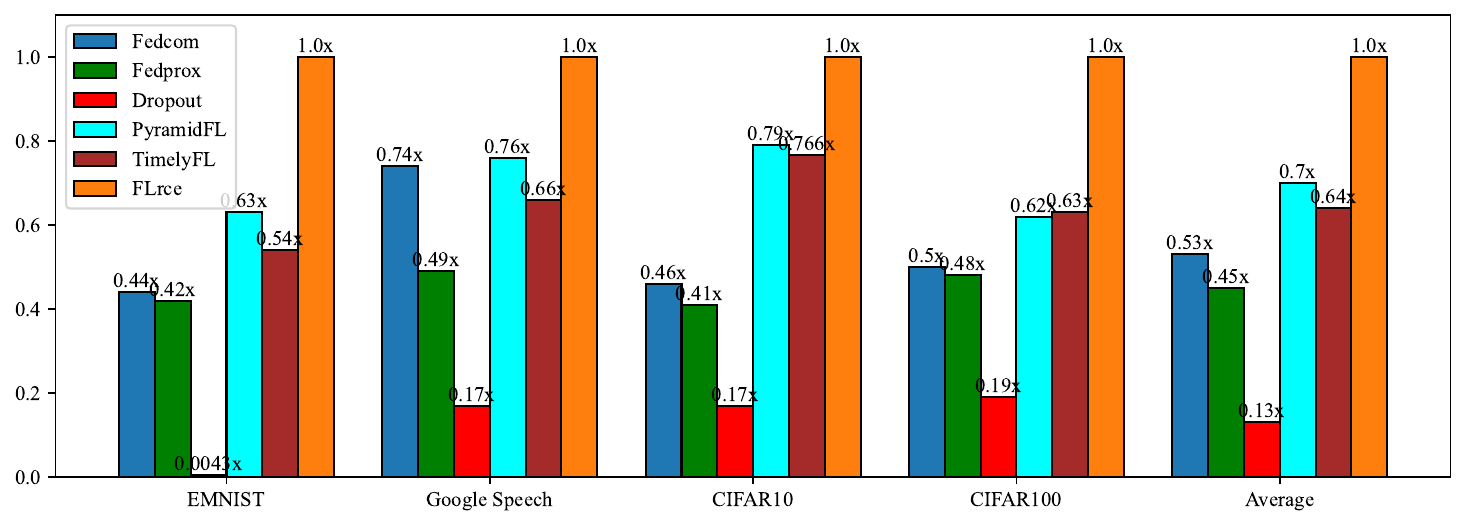}
    \caption{
    Computation efficiency.}
    \label{img:compeff}
\end{figure*}
\subsubsection{Computation efficiency}
We use energy consumption as a measurement of the computation cost \cite{flRCsurvey, autofl}. During the local training process, We measure the energy consumption of the Jetson Nano Developer Kits using a plug-in power monitor \cite{monitor}, when the training is completed, we sum up all clients' energy consumption to derive the overall computation cost of the entire system, the results are available in Figure \ref{img:energy}. As shown by Figure \ref{img:energy}, FLrce consumes the least energy on all datasets, demonstrating the strength of FLrce in reducing the computation costs. Furthermore, motivated by \cite{autofl}, we combine accuracy and energy consumption to explore the computation efficiency of the benchmarks. Given the final accuracy and energy consumption, the computation efficiency is given by Equation (\ref{eq:compeff}):
\begin{equation}\label{eq:compeff}
    \text{Computation efficiency} \triangleq \frac{\text{accuracy}}{\text{energy consumption}}
\end{equation}
As defined by Equation (\ref{eq:compeff}), the computation efficiency is proportional to accuracy and the inverse of energy consumption. It represents how a FL method is capable of utilizing the given computation resource to train an accurate global model. For a FL method, the higher accuracy it achieves and the less energy it consumes, the more computation efficient it will be. Figure \ref{img:compeff} shows the computation efficiency of the benchmarks, for better visualization, the results have been normalized. As shown by Figure \ref{img:compeff}, FLrce obtains the highest computation efficiency on all of the four datasets. On average, FLrce improves the relative computation efficiency by at least 30\%. This result demonstrates FLrce's outstanding capability of achieving remarkable performance with limited computation resources. \par
\begin{figure*}
    \begin{subfigure}[b]{0.24\textwidth}
        \centering
        \includegraphics[width=\textwidth,
        height=0.13\textheight]{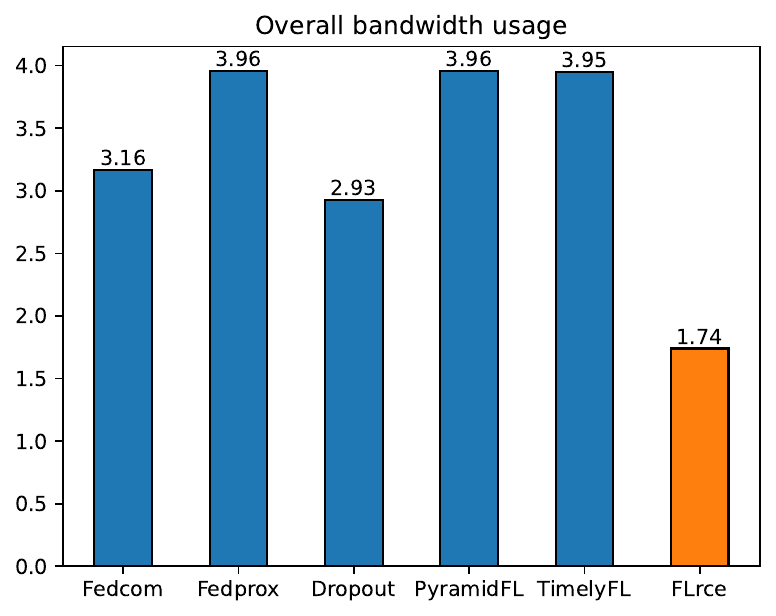}
        \caption{EMNIST}
        \label{emnistbandwidth}
    \end{subfigure}
    \begin{subfigure}[b]{0.24\textwidth}
        \centering
        \includegraphics[width=\textwidth,
        height=0.13\textheight]{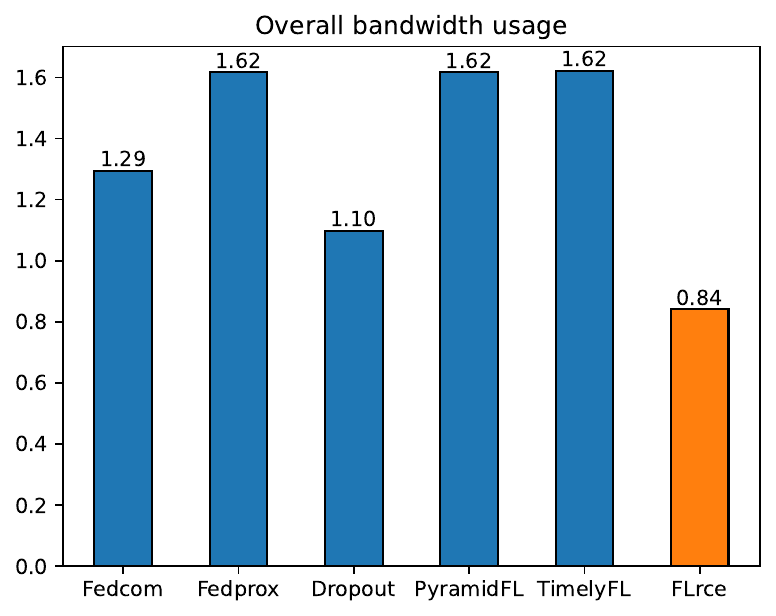}
        \caption{Google Speech}
        \label{voicebandwidth}
    \end{subfigure}
    \begin{subfigure}[b]{0.24\textwidth}
        \centering
        \includegraphics[width=\textwidth,
        height=0.13\textheight]{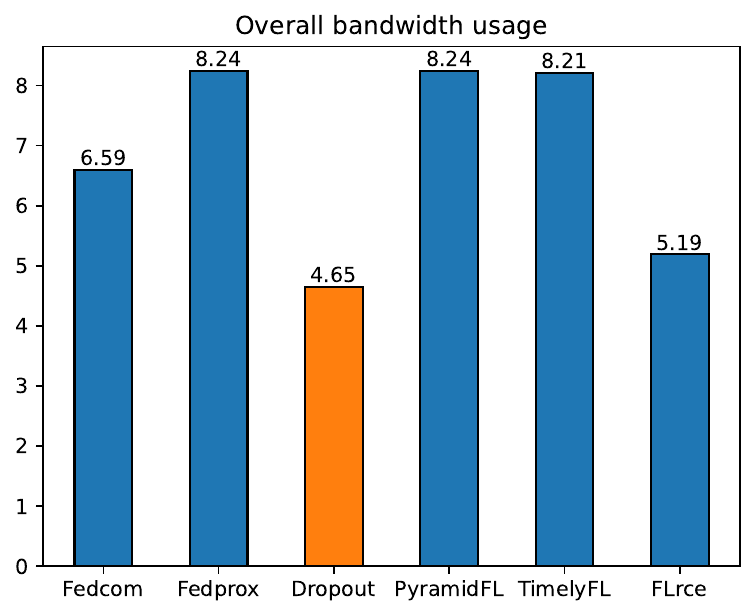}
        \caption{CIFAR10}
        \label{cifar10bandwidth}
    \end{subfigure}
    \begin{subfigure}[b]{0.24\textwidth}
        \centering
        \includegraphics[width=\textwidth, height=0.13\textheight]{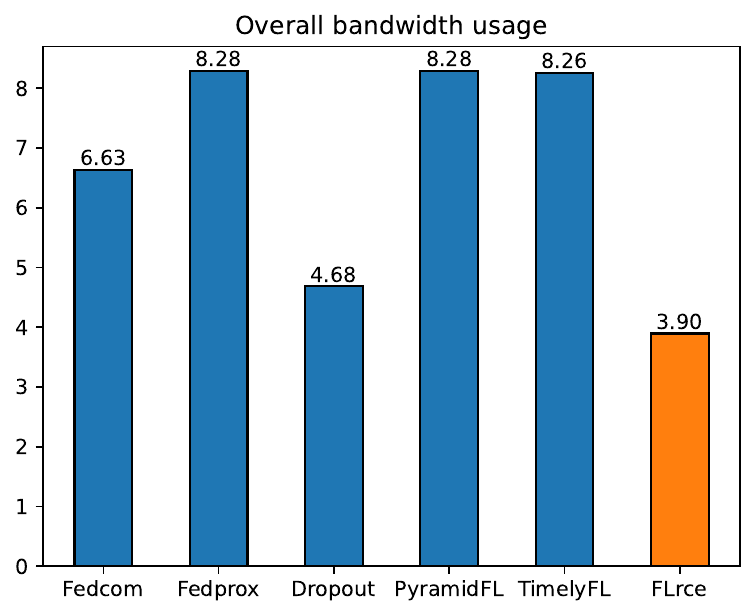}
        \caption{CIFAR100}
        \label{cifar100bandwidth}
    \end{subfigure}
    \caption{Overall bandwidth usage (GB).}
    \label{img:bandwidth}
\end{figure*}
\begin{figure*}
    \centering
    \includegraphics[width=\textwidth]{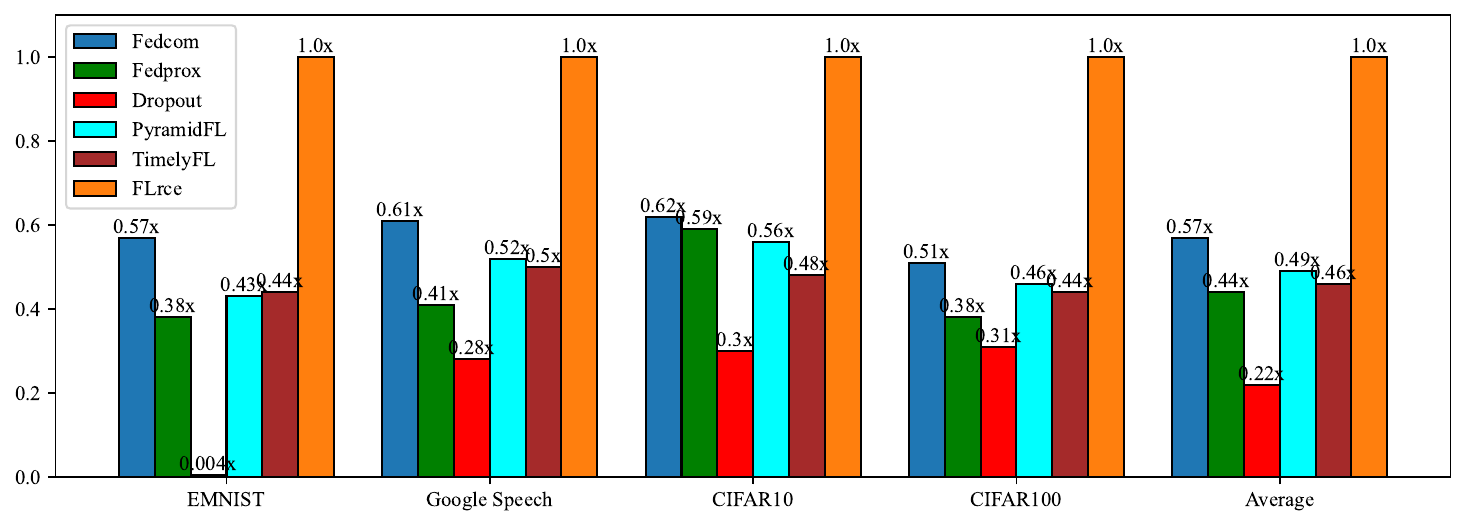}
    \caption{
    Communication efficiency.}
    \label{img:commeff}
\end{figure*}

\subsubsection{Communication efficiency}
We use bandwidth usage as a measurement of the communication cost, which is equivalent to the total size of exchanged parameters \cite{flstc, fedproto}. In the experiment, all the variables are implemented as the type of float32. Therefore, the size of transmitted messages equals 32 times the number of exchange parameters (with unit bytes). Figure \ref{img:commeff} lists the overall bandwidth usage of all benchmarks. Additionally, similar to computation efficiency, we define communication efficiency as Equation (\ref{eq:commeff}) shows. FL methods with higher accuracy and smaller bandwidth usage will have larger communication efficiency. 
\begin{equation}\label{eq:commeff}
    \text{Communication efficiency} \triangleq \frac{\text{accuracy}}{\text{bandwidth usage}}
\end{equation}
From Figure \ref{img:bandwidth}, we can see that FLrce has the least bandwidth usage on EMNIST, Google Speech and CIFAR100. For CIFAR10, FLrce has the second least bandwidth usage that is slightly higher than Dropout. Even though Dropout uses less bandwidth than FLrce on CIFAR10, the final accuracy of Dropout is much lower than FLrce (14.9\% vs. 57.8\%). In this case, Dropout does not achieve a high communication efficiency as it does not maintain accuracy. Figure \ref{img:commeff} shows that FLrce outperforms the baselines in terms of communication efficiency. On average, FLrce improves the relative communication efficiency by at least 43\%.

\subsection{Effect of the Early Stopping Threshold}
\begin{figure*}
    \begin{subfigure}[b]{0.24\textwidth}
        \centering
        \includegraphics[width=\textwidth,
        height=0.13\textheight]{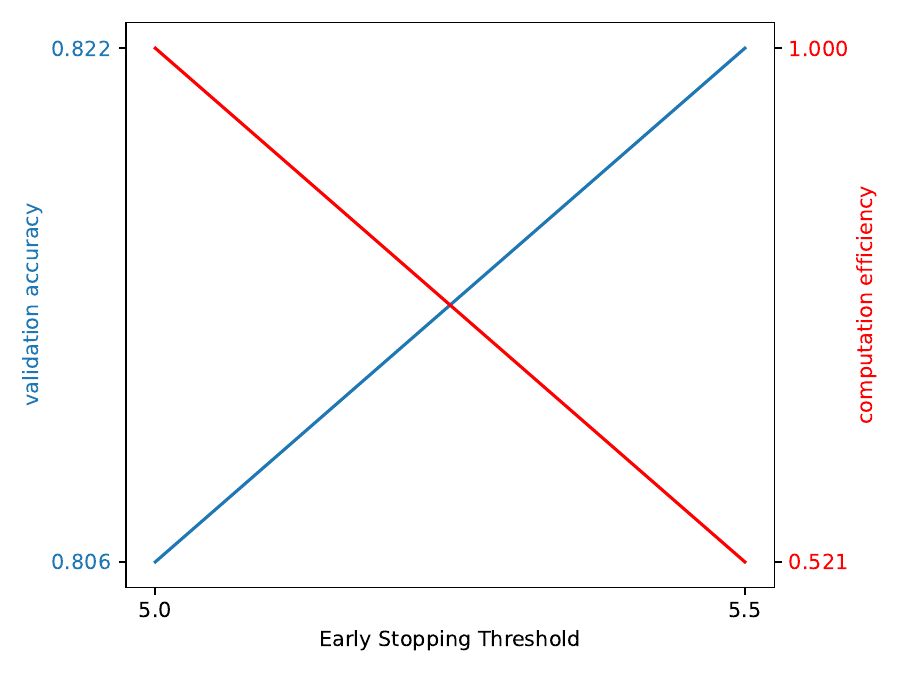}
        \caption{EMNIST}
        \label{emnistes_comp}
    \end{subfigure}
    \begin{subfigure}[b]{0.24\textwidth}
        \centering
        \includegraphics[width=\textwidth,
        height=0.13\textheight]{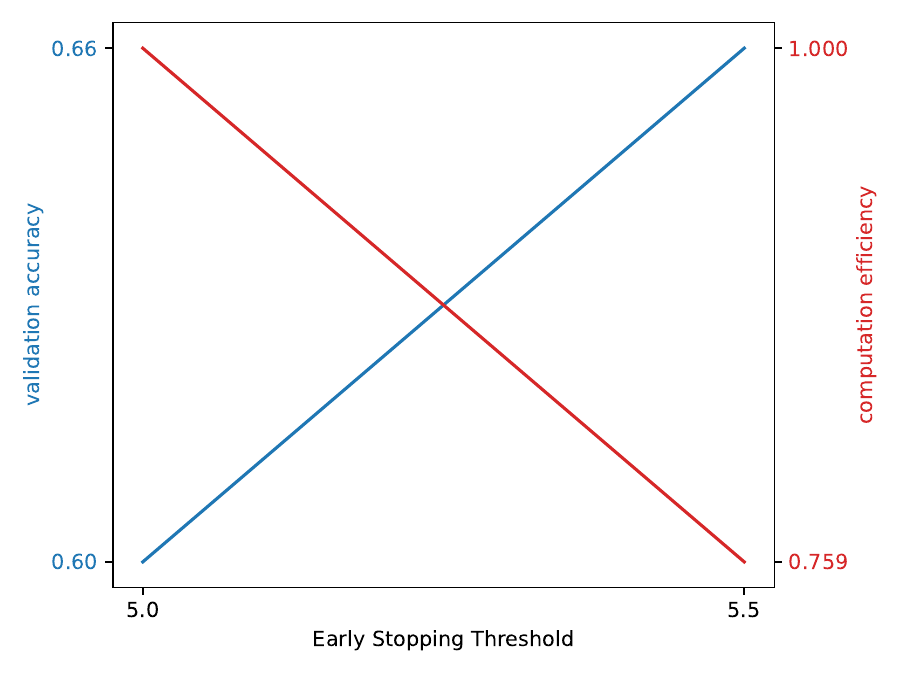}
        \caption{Google Speech}
        \label{voicees_comp}
    \end{subfigure}
    \begin{subfigure}[b]{0.24\textwidth}
        \centering
        \includegraphics[width=\textwidth,
        height=0.13\textheight]{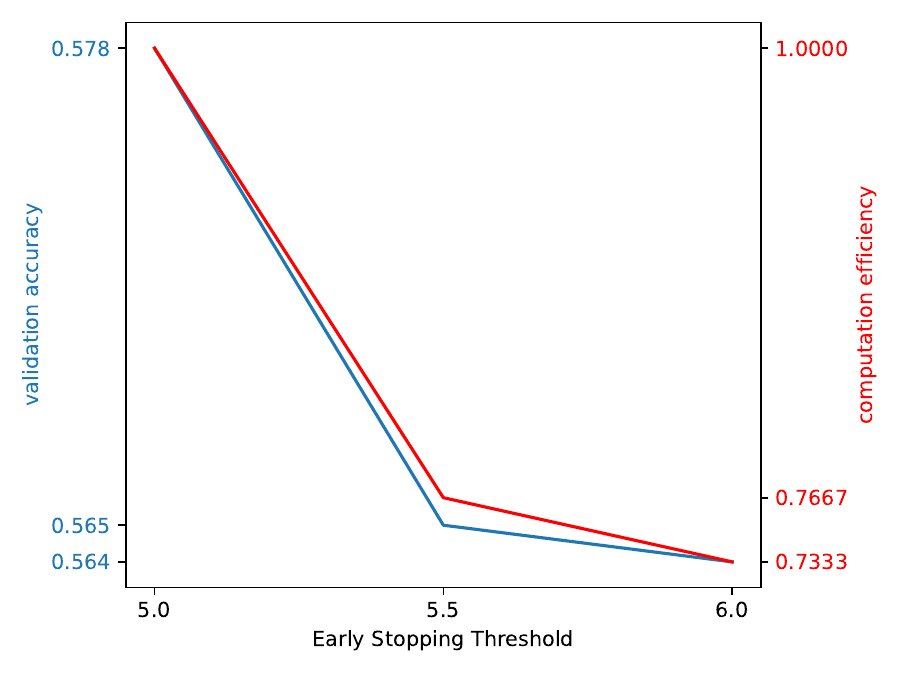}
        \caption{CIFAR10}
        \label{cifar10es_comp}
    \end{subfigure}
    \begin{subfigure}[b]{0.24\textwidth}
        \centering
        \includegraphics[width=\textwidth, height=0.13\textheight]{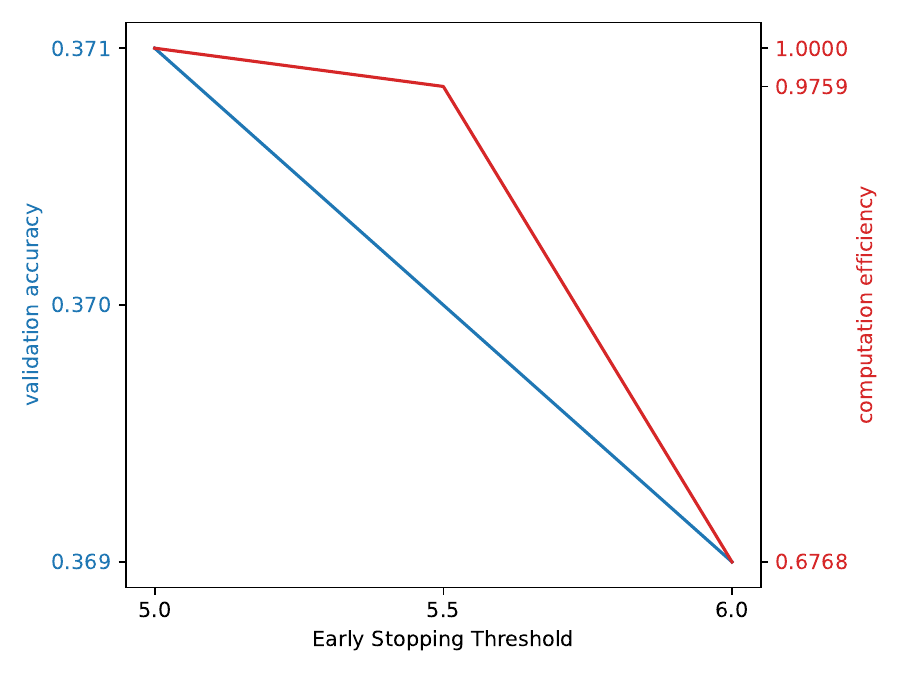}
        \caption{CIFAR100}
        \label{cifar100es_comp}
    \end{subfigure}
    \caption{Effect of the Early Stopping threshold on computation efficiency, the coordinates have been normalized for better visualization.}
    \label{img:esparameter_comp}
\end{figure*}

\begin{figure*}
    \begin{subfigure}[b]{0.24\textwidth}
        \centering
        \includegraphics[width=\textwidth,
        height=0.13\textheight]{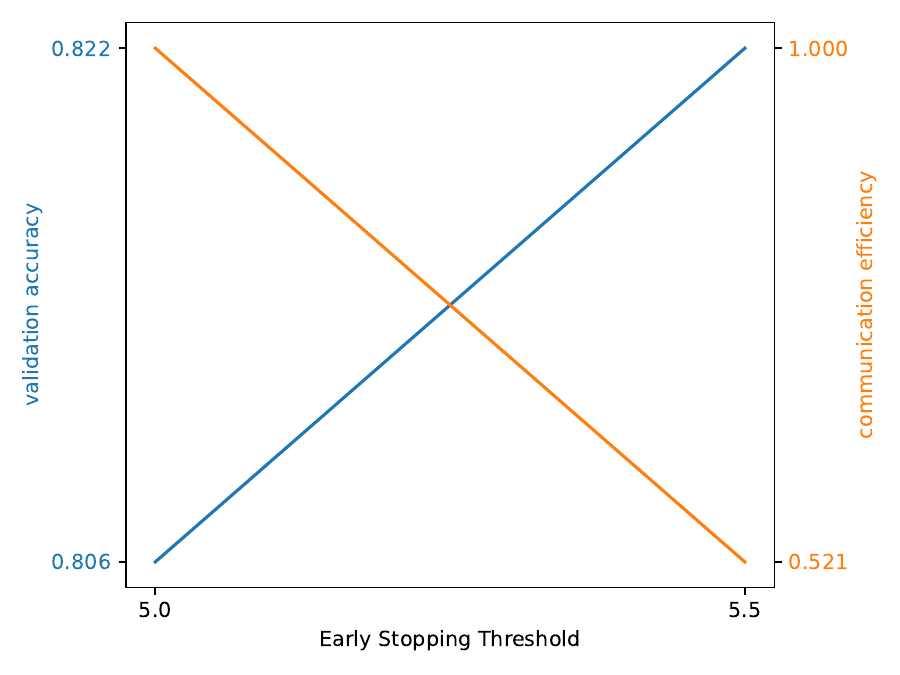}
        \caption{EMNIST}
        \label{emnistes}
    \end{subfigure}
    \begin{subfigure}[b]{0.24\textwidth}
        \centering
        \includegraphics[width=\textwidth,
        height=0.13\textheight]{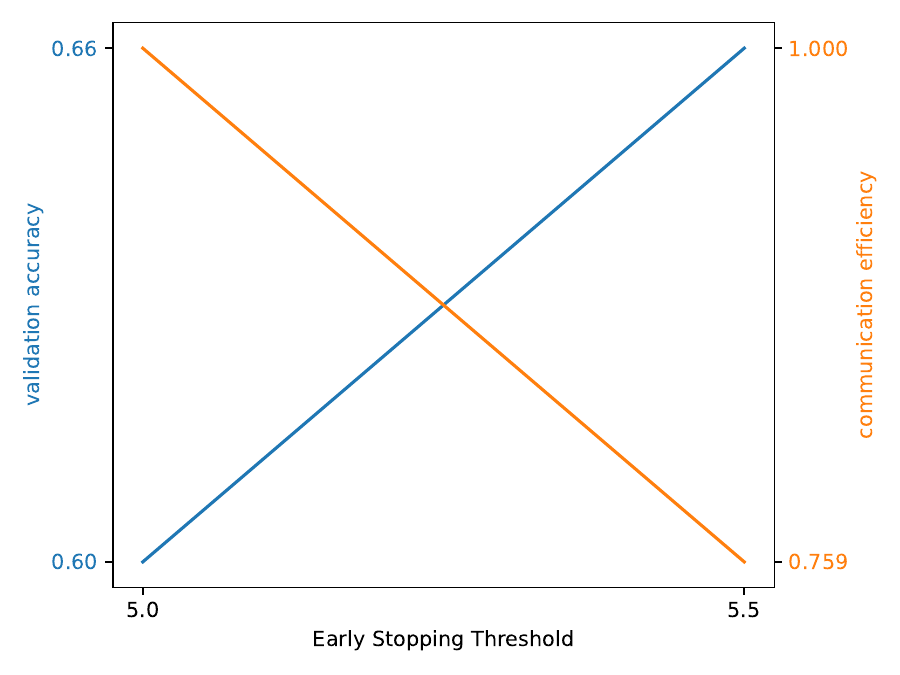}
        \caption{Google Speech}
        \label{voicees}
    \end{subfigure}
    \begin{subfigure}[b]{0.24\textwidth}
        \centering
        \includegraphics[width=\textwidth,
        height=0.13\textheight]{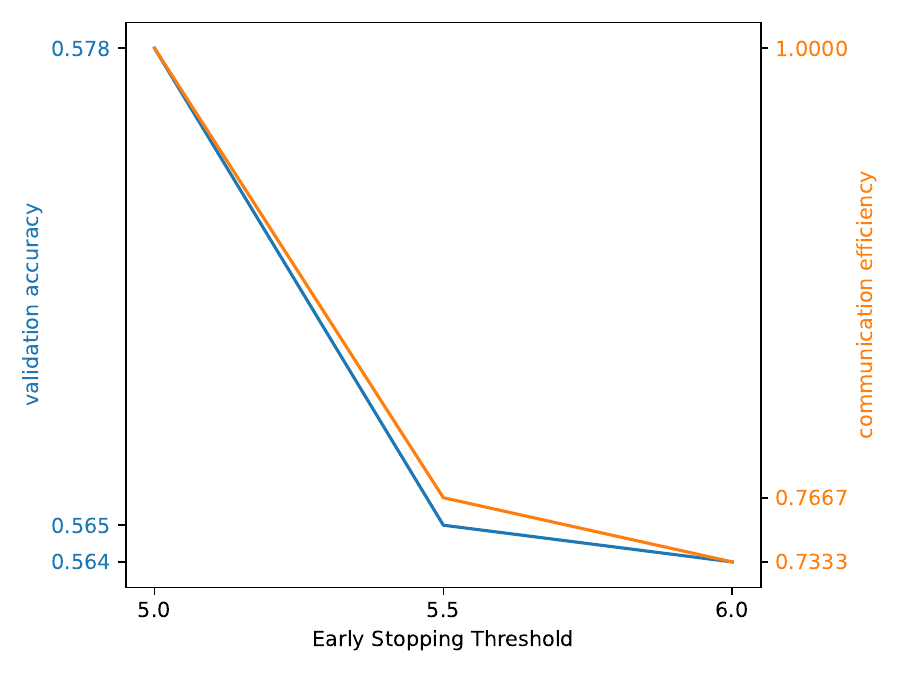}
        \caption{CIFAR10}
        \label{cifar10es}
    \end{subfigure}
    \begin{subfigure}[b]{0.24\textwidth}
        \centering
        \includegraphics[width=\textwidth, height=0.13\textheight]{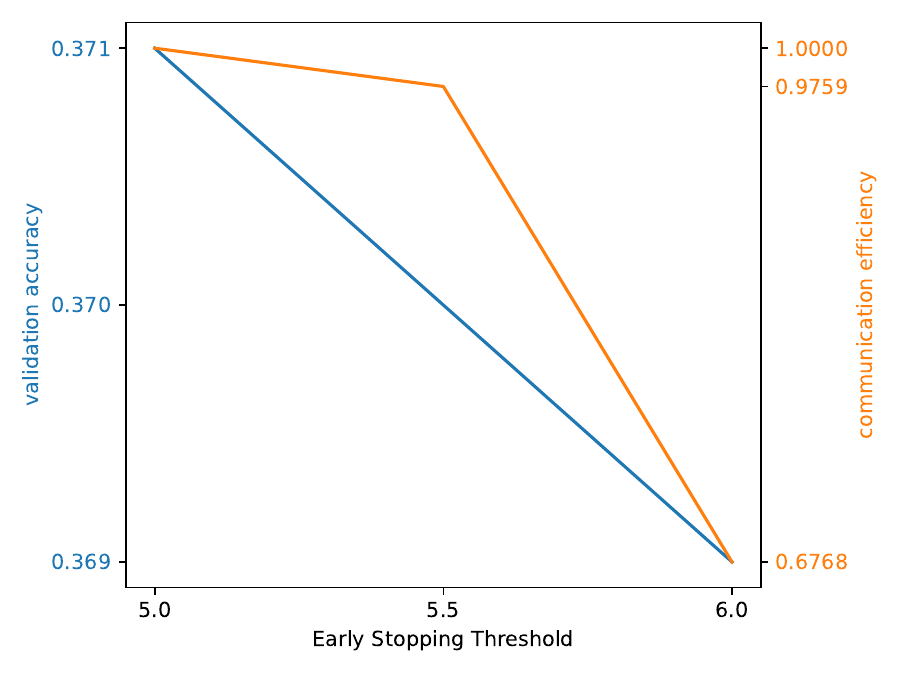}
        \caption{CIFAR100}
        \label{cifar100es}
    \end{subfigure}
    \caption{Effect of the Early Stopping threshold on communication efficiency, the coordinates have been normalized for better visualization.}
    \label{img:esparameter_comm}
\end{figure*}

We study the effect of the ES threshold \(\psi\) empirically. We run FLrce multiple times with different values of \(\psi\). As defined by subsection \ref{FLrce_es}, to make sure that FLrce properly stops at a point with an acceptable validation accuracy, the ES mechanism should only be activated when a relatively large degree of conflict occurs among the selected clients, that is, the value of \(\psi\) should not be too small. Furthermore, \(\psi\) should not be too large as well; otherwise, the ES mechanism might not be activated in time and cannot reduce the resource consumption. Therefore, we tune the values of \(\psi\) within \(\{4.0, 4.5, 5.0, 5.5, 6.0, 6.5, 7.0\}\) using the well-known grid search method, and the results are shown by Figure \ref{img:esparameter_comp}, Figure \ref{img:esparameter_comm}  and Table \ref{tab:es and rounds}. It is worth mentioning that the cases of \(\psi < 5\) and \(\psi > 6\) are not included in these results as they either terminate FLrce too early with low accuracy or fail to terminate before \(T\). Moreover, Figure \ref{img:esparameter_comp} and Figure \ref{img:esparameter_comm} look exactly the same. This is because the computation/communication cost of FLrce per round is expected to be the same, so the computation/communication efficiency is proportional to \(\frac{\text{accuracy}}{\text{round}}\) (see Equations (\ref{eq:compeff}) and (\ref{eq:commeff})). Therefore, after normalization, the relative values of the computation and communication efficiency both become \(\frac{\text{accuracy}}{\text{round}}\). For better illustration, we plot Figures \ref{img:esparameter_comp} and \ref{img:esparameter_comm} separately even though they show the same values. As shown by Table \ref{tab:es and rounds}, FLrce successfully stops on all Datasets with \(\psi=\) 5.0 or 5.5. For EMNIST and Google Speech, FLrce fails to stop before the maximum global iteration with \(\psi \geq\) 6. Therefore, to make sure the ES mechanism be successfully activated, \(\psi\) should be selected between \(5\sim5.5\). As shown by Figure \ref{img:esparameter_comp} and Figure \ref{img:esparameter_comm}, the value of \(\psi\) to achieve the highest computation/communication efficiency is 5.0 (i.e. \(\frac{P}{2}\), half of the number of selected clients per round). Accordingly, we set \(\psi\) to 5.0 for EMNIST, CIFAR10, CIFAR100. For Google Speech, we set \(\psi\) to 5.5 as it achieves higher accuracy than \(\psi=5.0\) while still achieving the highest efficiency compared with the baselines (see Figures \ref{img:compeff} and \ref{img:commeff}). Based on the empirical results, we suggest that resource-constrained IoT systems use a relatively small \(\psi\) (\(\approx\)50\% of the active clients \(P\)) to maximize the efficiency of utilizing the limited resources. On the other hand, IoT systems with sufficient resource budget can use a relatively large threshold (\(\approx\)55\(\sim\)60\% of the active clients \(P\)) to achieve higher model accuracy. We aim to find the optimal value of \(\psi\) via theoretical analysis in future works. \par
\begin{table*}
    \centering
    \begin{tabular}{c|cc|cc|cc}
    \hline
         \multirow{2}{10mm}{\textbf{Dataset}} & \multicolumn{2}{|c|}{\(\psi=5.0\)} & \multicolumn{2}{|c|}{\(\psi=5.5\)} & \multicolumn{2}{c}{\(\psi=6.0\)} \\
         &Accuracy&ES Round&Accuracy&ES Round&Accuracy&ES Round\\
         \hline
         EMNIST & 80.6\% & 44 & 82.2\% & 83 & N/A & N/A \\
         Google Speech & 60.5\% & 36 & 66.0\% & 52 & N/A & N/A\\
         CIFAR10 & 57.8\% & 63 & 56.5\% & 81 & 56.4\% & 85 \\
         CIFAR100 & 37.1\% & 47 & 37.0\% & 48 & 36.9\% & 69 \\
    \hline
    \end{tabular}
    \caption{Round and accuracy corresponding to different values of the Early Stopping threshold \(\psi\). Note that values of \(\psi\) that are too small (\(< 5\)) or too large (\(>6\)) are not shown in this table. For one thing, if \(\psi\) is too small, FLrce terminates in the very first few rounds with low final accuracy. For another thing, if \(\psi\) is too large, the ES mechanism cannot be triggered before the maximum iteration \(T\).}
    \label{tab:es and rounds}
\end{table*}
\subsection{Effect of the Early Stopping Mechanism} 
\begin{figure*}
    \begin{subfigure}[b]{0.24\textwidth}
        \centering
        \includegraphics[width=\textwidth,
        height=0.13\textheight]{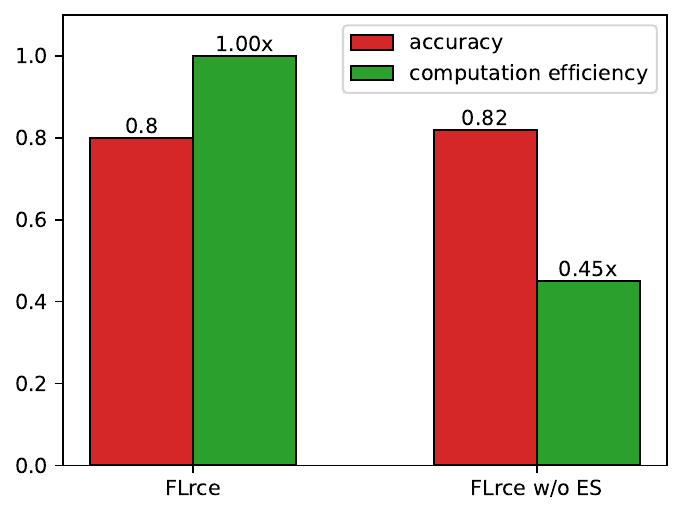}
        \caption{EMNIST}
        \label{emnistesornot_comp}
    \end{subfigure}
    \begin{subfigure}[b]{0.24\textwidth}
        \centering
        \includegraphics[width=\textwidth,
        height=0.13\textheight]{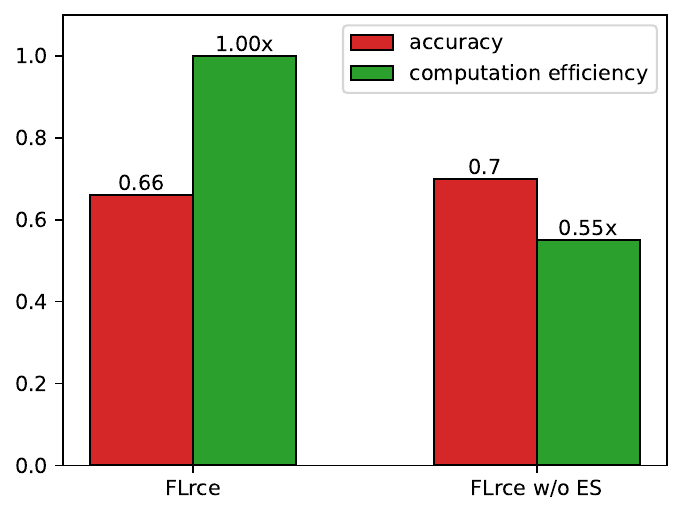}
        \caption{Google Speech}
        \label{voiceesornot_comp}
    \end{subfigure}
    \begin{subfigure}[b]{0.24\textwidth}
        \centering
        \includegraphics[width=\textwidth,
        height=0.13\textheight]{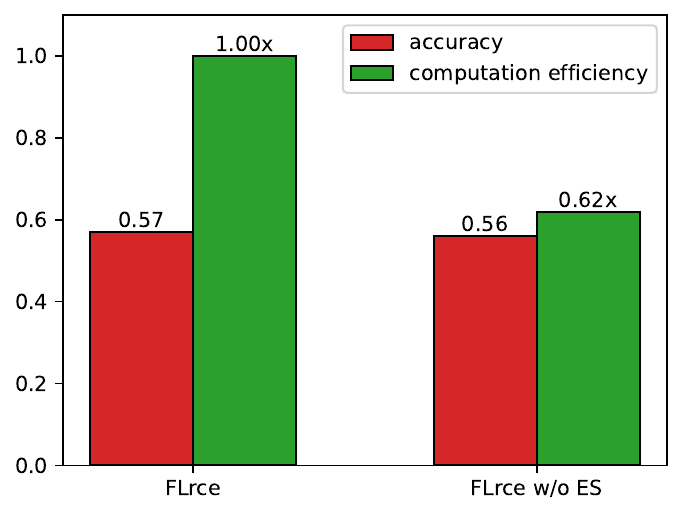}
        \caption{CIFAR10}
        \label{cifar10esornot_comp}
    \end{subfigure}
    \begin{subfigure}[b]{0.24\textwidth}
        \centering
        \includegraphics[width=\textwidth, height=0.13\textheight]{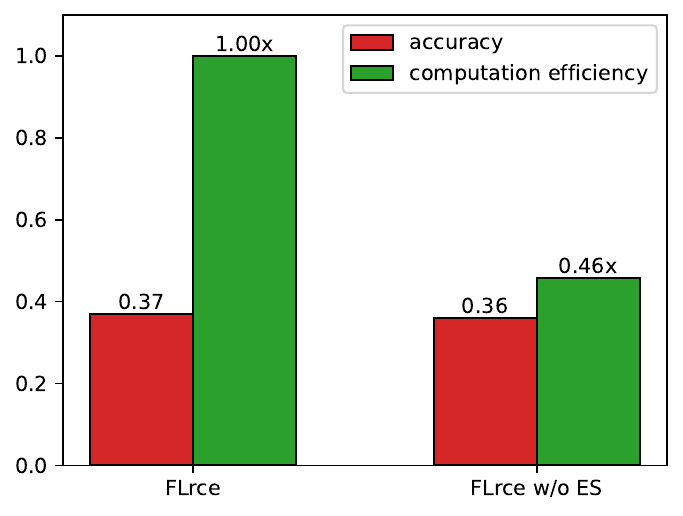}
        \caption{CIFAR100}
        \label{cifar100esornot_comp}
    \end{subfigure}
    \caption{Accuracy and computation efficiency of FLrce vs. FLrce without Early stopping.}
    \label{img:esornot_comp}
\end{figure*}

\begin{figure*}
    \begin{subfigure}[b]{0.24\textwidth}
        \centering
        \includegraphics[width=\textwidth,
        height=0.13\textheight]{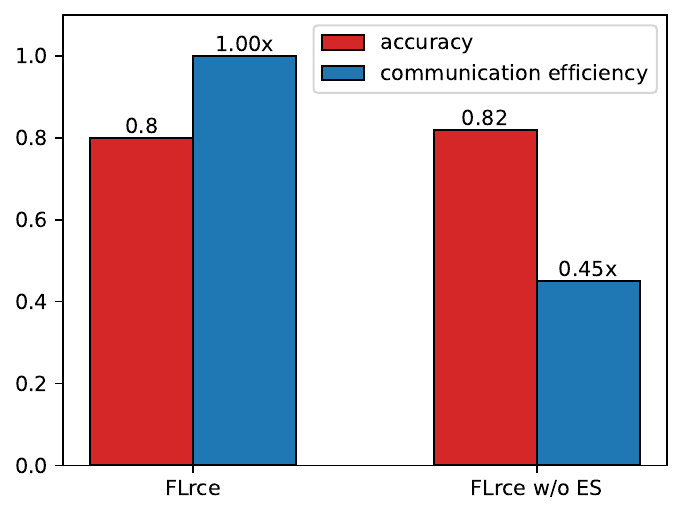}
        \caption{EMNIST}
        \label{emnistesornot_comm}
    \end{subfigure}
    \begin{subfigure}[b]{0.24\textwidth}
        \centering
        \includegraphics[width=\textwidth,
        height=0.13\textheight]{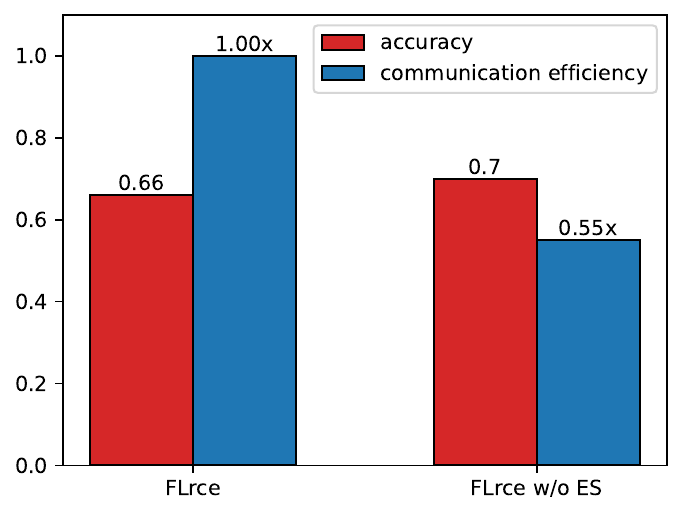}
        \caption{Google Speech}
        \label{voiceesornot_comm}
    \end{subfigure}
    \begin{subfigure}[b]{0.24\textwidth}
        \centering
        \includegraphics[width=\textwidth,
        height=0.13\textheight]{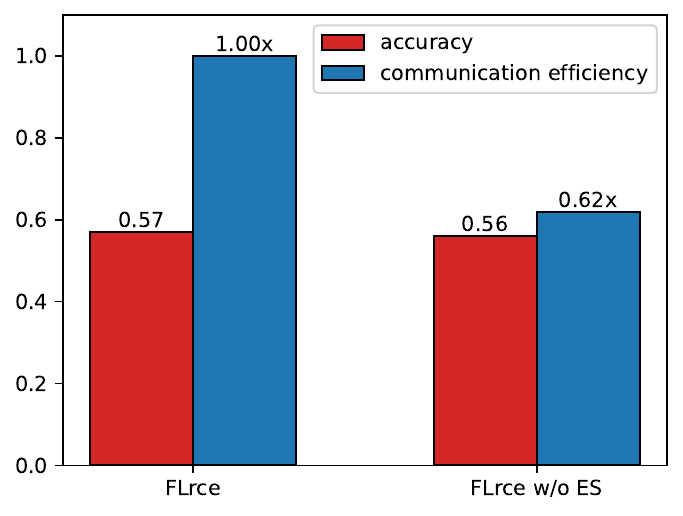}
        \caption{CIFAR10}
        \label{cifar10esornot_comm}
    \end{subfigure}
    \begin{subfigure}[b]{0.24\textwidth}
        \centering
        \includegraphics[width=\textwidth, height=0.13\textheight]{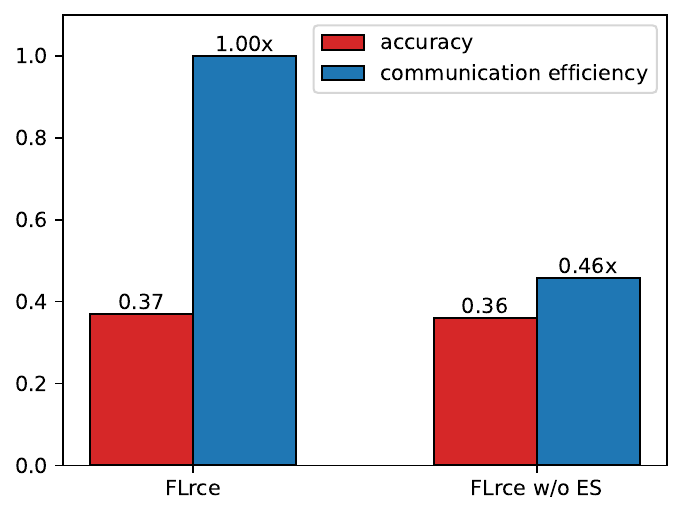}
        \caption{CIFAR100}
        \label{cifar100esornot_comm}
    \end{subfigure}
    \caption{Accuracy and communication efficiency of FLrce vs. FLrce without Early stopping.}
    \label{img:esornot_comm}
\end{figure*}

To investigate whether the ES mechanism will cause undesirable accuracy loss, we compare the final accuracy of FLrce and FLrce without early stopping (FLrce w/o ES), using the threshold \(\psi\) specified in the previous subsection. The only difference between FLrce and FLrce w/o ES is that the latter ignores the ES criterion and keeps training until reaching the maximum global iteration, and that relationship-based client selection strategy still applies in both settings. The comparison between FLrce and FLrce w/o ES is shown in Figure \ref{img:acc}, Figure \ref{img:esornot_comp} and Figure \ref{img:esornot_comm}. Note that Figures \ref{img:esornot_comp} and \ref{img:esornot_comm} present identical contents for the same reason as stated in the previous subsection. From Figures \ref{img:esornot_comp} and \ref{img:esornot_comm}, we can see that FLrce can significantly increase the efficiency of the resource utilization with marginal accuracy sacrifice. For EMNIST and Google Speech, FLrce degrades the final accuracy by 2\% and 4\% respectively compared with FLrce w/o ES, while the efficiency of FLrce w/o ES is only about 45\% and 55\% of FLrce. Besides, compared with existing efficient FL frameworks, FLrce still maintains competitive performance even with the accuracy deduction, as shown by Figure \ref{img:acc} and Table \ref{tab:acc and round}. For CIFAR10 and CIFAR100, FLrce even improves the final accuracy, as shown by Figures \ref{cifar10esornot_comp}, \ref{cifar10esornot_comm}, \ref{cifar100esornot_comp} and \ref{cifar100esornot_comm}. One possible reason is that FLrce prevents the over-fitting of FL by terminating the FL process before the global model converges to some local sub-optimal solution. \par

\subsection{Discussion}
\subsubsection{FLrce minimizes the accuracy loss caused by the trade-off between accuracy and efficiency}
From the experiment result, we can see that existing works usually make a "trade-off" strategy to realize efficiency, which might inevitably cause undesirable performance degradation. Specifically, in message compression methods, the server might not learn enough information from the partial parameter updates from clients, and might not be able to train an accurate global model. In accuracy relaxation methods, clients reduce the local training epoch to reduce computation costs, consequently, the acquired update information might not be enough to learn an accurate global model. In dropout methods, clients train and transmit sub-models to reduce computation and communication costs. However, dropping off parameters will modify the model architecture and make the sub-model learn inaccurate parameter update that diverges drastically from the original model. Figure \ref{img:acc} shows the validity of this analysis. \par
For comparison, FLrce tackles this bottleneck from a novel perspective, while existing works reduce the resource consumption by reducing the computation/communication cost per round, FLrce reduces the overall resource consumption by reducing the total rounds of training. The advantage of this method is that FLrce does not have to make any trade-off (compress gradients, less epoch, etc.) in the training process, and can achieve a high accuracy faster given the same training rounds as the baselines. Consequently, when FLrce terminates FL in advance, the accuracy loss will become marginal, as shown by Figures \ref{img:esornot_comp} and \ref{img:esornot_comm}. \par
\subsubsection{Independent implementation of the experiment methods}
As a matter of fact, different kinds of efficient FL frameworks are compatible with each other and can be combined to further improve efficiency. For instance, Dropout can further reduce the local epochs to save computation resources. FLrce can also transmit the compressed parameter updates to further reduce communication costs, or employ Dropout to overcome the computation/communication overheads of edge devices etc. To straightforwardly demonstrate the effectiveness of each method, the benchmarks are implemented independently in the experiment. We aim to find the optimal combination of the existing methods in future work. 
\subsubsection{Sub-model training does not necessarily reduce computation costs}
A point worth mentioning is that Dropout does not effectively reduce the computation cost as Figure \ref{img:energy} shows. The possible reason is that Dropout mainly reduces the \textit{width} (neurons) of a neural network rather than the depth (layers), while the training complexity mainly depends on the depth of the neural network. For example, if the model is trained with Pytorch, the computation cost mainly comes from the backward propagation process following the gradient graph that covers the entire neural network, while the size of the graph mainly depends on the depth of the neural network \cite{gradientgraph}. Therefore, Dropout does not effectively improve the computation efficiency. Comparatively, reducing the epoch/rounds of training is more helpful to reduce the computation cost, as shown by Figure \ref{img:energy}, Fedprox and FLrce consume less energy than Dropout.
\subsubsection{Generality of FLrce across various deep-learning tasks}
In the experiment, FLrce exhibits the ability of maintaining computation and communication consistently across different types of deep-learning tasks, including image classification (EMNIST, CIFAR10, CIFAR100) and speech recognition (Google Speech). Firstly, as shown in Figure 10, the learning curve of FLrce on Google Speech presents similar patterns to that on EMNIST, CIFAR10 and CIFAR100. Secondly, as shown in Figures 12 and 14, FLrce achieves the highest computation and communication efficiency in both image classification and speech recognition tasks. Lastly, as shown in Table 3, the early stopping points where FLrce terminates prematurely in speech recognition (52) and image classification (44\(\sim\)63, averaging 51.3) are close. Overall, due to the generality displayed in the experiment, we expect to explore the potentiality of FLrce in broad areas of IoT applications in the future.  
\section{related work} \label{RW}
In this section, we briefly introduce the relevant literature with respect to client heterogeneity and resource efficiency in FL. 
\subsection{Efficient Federated Learning}
To curtail the excessive communication and computation costs in FL, substantial studies have been conducted. To the best of our knowledge, the prior art can be divided into three categories: (1) \textit{message compression}, (2) \textit{accuracy relaxation}, and (3) \textit{dropout}. \textit{Message compression} methods reduce the communication cost by decreasing the volume of transmitted messages. For instance, in \cite{flexcom, flstc}, parameter updates are sparsified before communication, with the elements less than a particular threshold removed. \cite{qsgd} and \cite{terngrad} map the continuous elements of a gradient to discrete values without changing the statistical property. \cite{8bit} replaces the 32-bit gradients with 8-bit approximations. \cite{tinyscript} proposes a distribution-aware quantization framework that minimizes the quantization variance of gradients to mitigate accuracy sacrifice. Unfortunately, message compression methods do not consider the shortage of computation resources. \textit{Accuracy relaxation methods} save computation resources by allocating simpler tasks to clients. For example, \cite{fedprox}, \cite{fedparl} and \cite{pyramid} allow clients to train a sub-optimal solution during local training, which requires less computation effort. However, the communication resource shortage is neglected in this kind of approach. \textit{Dropout} simultaneously realizes communication and computation efficiency by allowing clients to train and transmit light-sized sub-models. To extract a sub-model from the global model, \cite{feddrop, randdrop} randomly prune neurons in the original neural network. \cite{fjord} and \cite{heterofl} prune neurons in a fixed order. In \cite{hermes}, each client prunes the unimportant neurons personally, where the unimportant neurons are determined by a group-lasso-based regularization. \par
Even though these works have made remarkable progress in saving computation/communication resources in FL, they lack a mechanism that evaluates the importance of FL clients. In real-world IoT applications, these works' performance might be compromised by the heterogeneous training contributions among clients. \par
\subsection{Federated Learning with Heterogeneous Client Contributions}
Due to the intrinsic non-iid local data distributions in IoT, clients contribute unequally to the FL task. Some clients promote FL by contributing important parameter updates, while others may hinder FL by sending malicious or noisy updates. Several studies have been conducted on how to maintain FL's performance with heterogeneous client training contributions. As mentioned before, existing works mainly adopt two ways. One way is to \textit{mitigate the negative effects of malicious updates in the aggregation phase}. For example, \cite{fltrust} designs a trust mechanism that assigns higher weights to clients whose gradients are more consistent with the global update. \cite{adbfl} designs an anomaly-detection-based FL framework that can identify malicious updates through a variation auto-encoder (VAE). All updates are fed into the VAE before aggregation, and those with a large construction error will be considered malicious and removed from the aggregation. \cite{robustfed} evaluates the reliability of clients using the truth inference technique in crowdsourcing and assigns lower weights to clients with low reliability in aggregation. The other way is to \textit{identify and discard malicious clients during the phase of client selection.} \cite{fldetecrtor} detects malicious clients based on update inconsistency to defend FL against large-scale model poisoning attacks. \cite{fedcbs} invents a metric named Quadratic Class-Imbalance Degree (QCID) that measures the distance between clients' local data distribution to an ideal uniform distribution. By selecting clients with the lowest QCID values, \cite{fedcbs} can improve the model's accuracy when the class distributions among clients are unbalanced. \par
Unfortunately, these works do not take the resource shortage issue into account, and subsequently become prohibitive in resource-constrained IoT systems. \par 
Overall, none of the aforementioned works can simultaneously address the problems of client heterogeneity and resource shortage. Comparatively, FLrce is able to reduce the overall consumption of computation and communication resources, meanwhile maintaining performance with unbalanced contributions from heterogeneous clients. \par
\section{Conclusion} \label{conclusion}

In this paper, we propose FLrce, an efficient FL framework excelling in overall communication and computation efficiency while maintaining accuracy with heterogeneous client contributions. Experiment results show that FLrce achieves remarkable accuracy under unbalanced data distributions and significantly improves the efficiency of communication and computation resources utilization. \par
The major limitation of FLrce is that it conducts full-model training on all clients regardless of the resource constraints of their edge devices. For devices that cannot train the entire global model because of limited capacity, this might become a bottleneck. As a solution, in the future, we aim to enhance the adaptability of FLrce to clients with limited device capacity. For example, consolidating FLrce with model pruning \cite{randdrop} and layer freezing \cite{cocofl} techniques to deal with resource-constrained devices’ incapability of training the entire global model. \par 
In addition, we plan to investigate how to improve the efficiency for individual clients, e.g., by leveraging multi-task optimization \cite{song2019multitasking}, how to explore the relationships between clients more effectively, e.g., via learning vector quantization techniques \cite{qin2005initialization}, and conduct an in-depth theoretical study on FLrce.
\section*{Acknowledgement}
This work is funded by the Australian Research Council under Grant No. DP220101823, DP200102611, and LP180100114.


\ifCLASSOPTIONcaptionsoff
  \newpage
\fi

\bibliography{reference.bib}{}
\bibliographystyle{IEEEtran}

\begin{IEEEbiography}[{\includegraphics[width=1in,height=1.25in,clip,keepaspectratio]{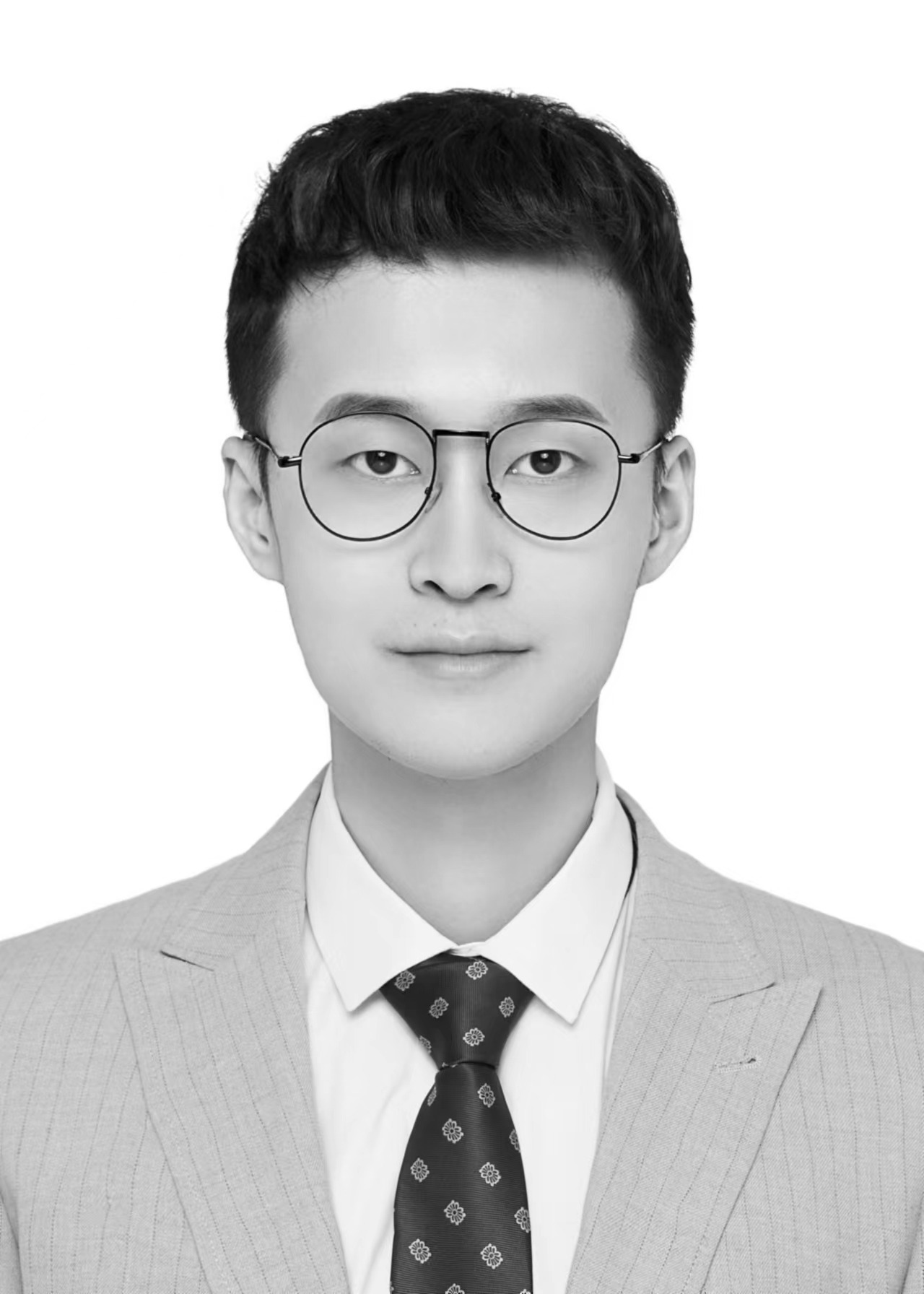}}]{Ziru Niu}
received the Bachelor degree and Master degree in 2019 and 2021 respectively, both from the University of Melbourne, VIC, Australia. He is currently a PhD candidate from the School of Computing Technologies, RMIT University, Melbourne, VIC, Australia. His research interests include machine learning, federated learning, edge computing, and the Internet of Things.
\end{IEEEbiography}

\begin{IEEEbiography}[{\includegraphics[width=1in,height=1.25in,clip,keepaspectratio]{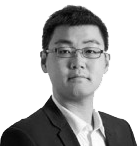}}]{Hai Dong} (Senior Member, IEEE) 
received a Ph.D. degree from Curtin University, Perth, Australia, and a Bachelor's degree from Northeastern University, Shenyang, China. He is currently a Senior Lecturer at the School of Computing Technologies, RMIT University, Melbourne, Australia. He also serves as a Chair for the IEEE Computational Intelligence Society Task Force on Deep Edge Intelligence. His primary research interests include Service-Oriented Computing, Edge Intelligence, Blockchain, Cyber Security, and Machine Learning. His publications appeared in ACM Computing Surveys, IEEE Transactions on Industrial Electronics, IEEE Transactions on Information Forensics and Security, IEEE Transactions on Mobile Computing, IEEE Transactions on Services Computing, IEEE Transactions on Software Engineering, ASE, ICML, etc. 
\end{IEEEbiography}

\begin{IEEEbiography}[{\includegraphics[width=1in,height=1.25in,clip,keepaspectratio]{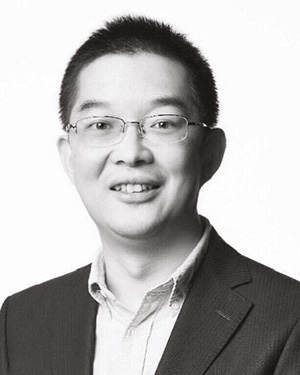}}]{A. K. Qin} (Senior Member, IEEE) received the B.Eng. degree in Automatic Control from Southeast University, Nanjing, China, in 2001, and the Ph.D. degree in Computer Science and Engineering from Nanyang Technology University, Singapore, in 2007. From 2007 to 2017, he was with the University of Waterloo, Waterloo, ON, Canada; INRIA Grenoble Rhône-Alpes, Montbonnot-Saint-Martin, France; and RMIT University, Melbourne, VIC, Australia. He joined Swinburne University of Technology, Hawthorn, VIC, Australia, in 2017, where he is now a Professor. He is currently the Director of Swinburne Intelligent Data Analytics Lab and the Deputy Director of Swinburne Space Technology and Industry Institute. His major research interests include machine learning, evolutionary computation, computer vision, remote sensing, services computing, and edge computing. 

Dr. Qin was a recipient of the 2012 IEEE Transactions ON Evolutionary Computation Outstanding Paper Award and the 2022 IEEE Transactions ON Neural Networks AND Learning Systems Outstanding Associate Editor. He is currently the Chair of the IEEE Computational Intelligence Society (CIS) Neural Networks Task Force on “Deep Vision in Space”, the Vice-Chair of the IEEE CIS Emergent Technologies Task Force on “Multitask Learning and Multitask Optimization”, the Vice-Chair of the IEEE CIS Neural Networks Task Force on “Deep Edge Intelligence”. He served as the General Co-Chair of the 2022 IEEE International Joint Conference on Neural Networks (IJCNN 2022) and as the Chair of the IEEE CIS Neural Networks Technical Committee during the 2021-2022 term.

\end{IEEEbiography}

\begin{IEEEbiography}[{\includegraphics[width=1in,height=1.25in,clip,keepaspectratio]{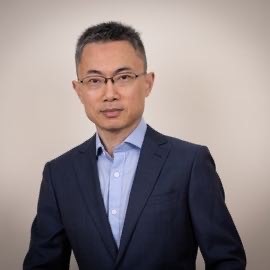}}]{Tao Gu} (Fellow, IEEE) is currently a Professor in Department
of Computing at Macquarie University, Sydney.
He obtained his Ph.D. in Computer Science from
National University of Singapore, M.Sc. in Electrical and Electronic Engineering from Nanyang
Technological University, and B.Eng. in Automatic Control from Huazhong University of Science and Technology. His current research interests include Internet of Things, Ubiquitous Computing, Mobile Computing, Embedded AI, Wireless Sensor Networks, and Big Data Analytics. He is a Distinguished Member of the ACM.
\end{IEEEbiography}

\end{document}